%% file: main.tex
\documentclass{article}

     \usepackage[final]{neurips_2021}

\usepackage[utf8]{inputenc} 
\usepackage[T1]{fontenc}    
\usepackage{url}            
\usepackage{booktabs}       
\usepackage{amsfonts}       
\usepackage{nicefrac}       
\usepackage{microtype}      
\usepackage{amsmath}
\usepackage{subcaption}
\usepackage{float}
\usepackage{makecell}
\usepackage{graphicx}
\usepackage{sidecap}
\usepackage{enumitem}
\usepackage{arydshln} 
\usepackage{amsthm}

\usepackage[usenames,dvipsnames]{xcolor}
\definecolor{shadecolor}{gray}{0.9}

\newcommand{\red}[1]{\textcolor{BrickRed}{#1}}
\newcommand{\orange}[1]{\textcolor{BurntOrange}{#1}}
\newcommand{\green}[1]{\textcolor{OliveGreen}{#1}}
\newcommand{\blue}[1]{\textcolor{Blue}{#1}}
\newcommand{\purple}[1]{\textcolor{Plum}{#1}}

\newcommand{\gold}[1]{\textcolor{Goldenrod}{#1}}

\usepackage{natbib}
\usepackage[colorlinks,linktoc=all]{hyperref}
\usepackage[all]{hypcap}
\definecolor{mydarkblue}{rgb}{0,0.08,0.45} 
\hypersetup{citecolor=mydarkblue}
\hypersetup{linkcolor=mydarkblue}
\hypersetup{urlcolor=mydarkblue}

\usepackage[nameinlink]{cleveref}
\crefname{section}{§}{§§}
\Crefname{section}{§}{§§}
\creflabelformat{equation}{#2\textup{#1}#3}  

\usepackage[textsize=scriptsize,textwidth=2.2cm]{todonotes}
\usepackage{etoolbox}
\newtoggle{showtodos}
\toggletrue{showtodos}

\input{math_commands}

\iftoggle{showtodos}{
\newcommand{\mcm}[1]{\todo[color=magenta!20]{Mike: #1}}
\newcommand{\archit}[1]{\todo[color=green!40]{Archit: #1}}
\newcommand{\nick}[1]{\todo[color=blue!40]{Nick: #1}}
\newcommand{\dustin}[1]{\todo[color=purple!20]{Dustin: #1}}
\newcommand{\balaji}[1]{\todo[color=red!20]{Balaji: #1}}
\newcommand{\becca}[1]{\todo[color=pink!20]{Becca: #1}}
}{
\newcommand{\mcm}[1]{}
\newcommand{\archit}[1]{}
\newcommand{\nick}[1]{}
\newcommand{\dustin}[1]{}
\newcommand{\balaji}[1]{}
\newcommand{\becca}[1]{}
}

\newtheorem{theorem}{Theorem}[section]
\newtheorem{proposition}[theorem]{Proposition}

\sidecaptionvpos{figure}{c}

\title{
Soft Calibration Objectives for Neural Networks
}

\author{
 Archit Karandikar\thanks{co-first author} \\
 Google Research\\
 \texttt{archk@google.com} \\
   \And
   Nicholas Cain$^*$ \\
   Google Research \\
   \texttt{nicholascain@google.com} \\
   \AND
   Dustin Tran \\
   Google Research \\
   \texttt{trandustin@google.com} \\
   \And
   Balaji Lakshminarayanan \\
   Google Research \\
   \texttt{balajiln@google.com} \\
   \And
   Jonathon Shlens \\
   Google Research \\
   \texttt{shlens@google.com} \\
   \AND
   Michael C.~Mozer \\
   Google Research \\
   \texttt{mcmozer@google.com} \\
   \And
   Becca Roelofs \\
   Google Research \\
   \texttt{rolfs@google.com} \\
}

\begin{document}

\maketitle

\begin{abstract}
Optimal decision making requires that classifiers produce uncertainty estimates consistent with their empirical accuracy. However, deep neural networks are often under- or over-confident in their predictions. Consequently, methods have been developed to improve the calibration of their predictive uncertainty, both during training and post-hoc. In this work, we propose differentiable losses to improve calibration based on a soft (continuous) version of the binning operation underlying popular calibration-error estimators. When incorporated into training, these soft calibration losses achieve state-of-the-art single-model ECE across multiple datasets with less than 1\% decrease in accuracy. For instance, we observe an 82\% reduction in ECE (70\% relative to the post-hoc rescaled ECE) in exchange for a 0.7\% relative decrease in accuracy relative to the cross-entropy baseline on CIFAR-100.
When incorporated post-training, the soft-binning-based calibration error objective improves upon temperature scaling, a popular recalibration method. 
Overall, experiments across losses and datasets demonstrate that using calibration-sensitive procedures yield better uncertainty estimates under dataset shift than the standard practice of using a cross-entropy loss and post-hoc recalibration methods.%
\footnote{Code available on GitHub: \href{https://github.com/google/uncertainty-baselines/tree/main/experimental/caltrain}{https://github.com/google/uncertainty-baselines/tree/main/experimental/caltrain}}

\end{abstract}

\section{Introduction}
\label{sec:introduction}
Despite the success of deep neural networks across a variety of domains, they are still susceptible to miscalibrated predictions. Both over- and under-confidence contribute to miscalibration, and empirically, deep neural networks empirically exhibit significant miscalibration \citep{Guo2017OnCO}. Calibration error (\emph{CE}) quantifies a model's miscalibration by measuring how much its confidence, i.e. the predicted probability of correctness, diverges from its accuracy, i.e. the empirical probability of correctness. Models with low CE are critical in domains where satisfactory outcomes depend on well-modeled uncertainty, such as autonomous vehicle navigation \citep{Bojarski2016EndTE} and medical diagnostics \citep{Jiang2012CalibratingPM, Caruana2015IntelligibleMF, Kocbek2020LocalIO}. Calibration has also been shown to be useful for improving model fairness \citep{Pleiss2017FairnessCalibration} and detecting out-of-distribution data \citep{Kuleshov2017EstimatingUO, Devries2018LearningCF, Shao2020CalibratingDN}. More generally, low CE is desirable in any setting in which thresholds are applied to the predicted confidence of a neural network in order to make a decision.



Methods for quantifying CE typically involve binning model predictions based on their confidence. 
CE is then computed empirically as a weighted average of the absolute difference in average prediction confidence and average accuracy across different bins \citep{Naeini2015ObtainingWC}. Oftentimes these bins are selected heuristically such as \emph{equal-width} (uniformly binning the score interval) and \emph{equal-mass} (with equal numbers of samples per-bin) \citep{nixon2019measuring}.

However, these commonly used measures of CE are not trainable with gradient-based methods because the binning operation is discrete and has zero derivatives. As a result, neural network parameters are not directly trained to minimize CE, either during training or during post-hoc recalibration. In this paper, we introduce new objectives based on a differentiable binning scheme that can be used to efficiently and directly optimize for calibration. 

%
\textbf{Contributions.} We propose estimating CE with soft (i.e., overlapping, continuous) bins rather than the conventional hard (i.e., nonoverlapping, all-or-none) bins. 
With this formulation, the CE estimate is differentiable, allowing us to use it as
(1) a secondary (i.e., auxiliary) loss to incentivize model calibration during training, and (2) a primary 
loss for optimizing post-hoc recalibration methods such as temperature scaling. 
In the same spirit, we soften the AvUC loss \citep{Krishnan2020ImprovingMC}, allowing us to use it as an effective secondary loss during training for non-Bayesian neural networks where the AvUC loss originally proposed for Stochastic Variational Inference (SVI) typically does not work. Even when training with the cross-entropy loss results in training set memorization (perfect train accuracy and calibration), Soft Calibration Objectives are still useful as secondary training losses for reducing test ECE using a procedure we call \emph{interleaved training}.

In an extensive empirical evaluation, we compare Soft Calibration Objectives as secondary losses to existing calibration-incentivizing losses. In the process, we find that soft-calibration losses outperform prior work on in-distribution test sets. Under distribution shift, we find that calibration-sensitive training objectives as a whole (not always the ones we propose) result in better uncertainty estimates compared to the standard cross-entropy loss coupled with temperature scaling. 

\begin{SCfigure}
   \centering
   \caption{We compare the effect size of various secondary losses on ECE (equal-mass binning, $\ell_2$ norm) across datasets and primary losses, both with and without post-hoc temperature scaling. The \red{S-AvUC} secondary loss we propose shows the strongest positive effect, followed by \purple{MMCE} and \orange{SB-ECE}. Note that a d-value of 0.8 (resp., 2.0) is considered a large (resp., huge) positive effect and d-values obtained here are much larger. Secondary losses which incentivize calibration show strong positive effect for reducing ECE even after temperature scaling.}
   \includegraphics[width=0.6\textwidth]{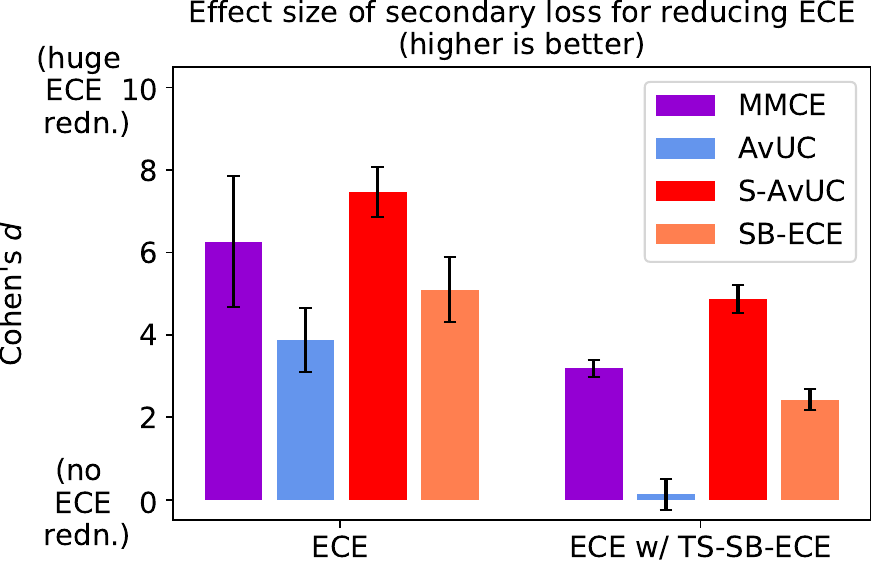}
   \label{fig:result-overview-cohen-d}
\end{SCfigure}

\begin{figure}[bt!]
   \centering
    \includegraphics[width=\textwidth]{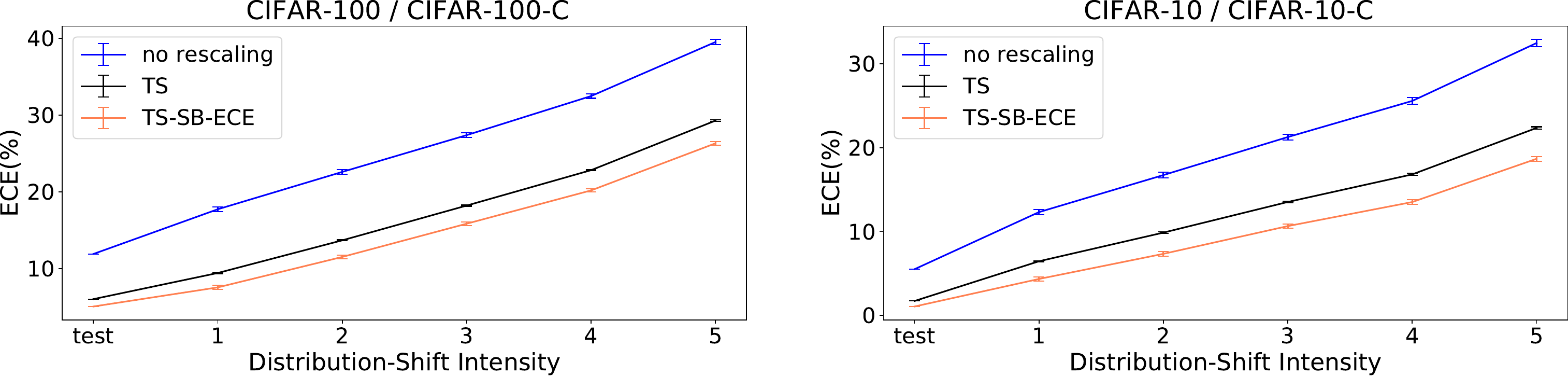}
  \caption{
  Post-hoc temperature scaling with the soft calibration error objective (\orange{TS-SB-ECE}) reduces ECE more than standard post-hoc temperature scaling (TS), particularly under distribution shift.
  This result holds across datasets (left and right panels),
  distribution shift intensities 
  (along abscissa) and training objectives (not shown). The training objective shown here for both datasets is the most popular one: NLL. The ECE value (equal-mass binning, $\ell_2$ norm) shown is the mean ECE across the corruption types that constitute CIFAR-10-C and CIFAR-100-C. Error bars are
  $\pm1$ standard error of mean (SEM), corrected for intrinsic variability due to type of corruption \citep{MassonLoftus}.}
  \vspace{-1em}
  \label{fig:ts-comparison}
\end{figure}

Our contributions can be summarized as follows:
\begin{itemize}
    \item We propose simple Soft Calibration Objectives S-AvUC, SB-ECE as secondary losses which optimize for CE
    \textit{throughout training}. We show that across datasets and choice of primary losses, the S-AvUC secondary loss results in the largest improvement in ECE as per the Cohen's $d$ effect-size metric (Figure \ref{fig:result-overview-cohen-d}). We also show that such composite losses obtain \textit{state-of-the-art single-model ECE} in exchange for less than 1\% reduction in accuracy (Figure \ref{fig:result-overview}) for CIFAR-10, CIFAR-100, and Imagenet.
    \item We improve upon temperature scaling - a popular post-hoc recalibration method - by directly optimizing the temperature parameter for soft calibration error instead of the typical log-likelihood.
    Our extension (TS-SB-ECE) consistently beats original temperature scaling (TS) across different datasets, loss functions and calibration error measures, and we find that the performance is better under dataset shift (Figure \ref{fig:ts-comparison}).
    \item Overall, our work demonstrates a fundamental advantage of objectives which better incentivize calibration over the standard practice of training with cross-entropy loss and then applying post-hoc methods such as temperature scaling. Uncertainty estimates of neural networks trained using these methods
    generalize better in and out-of-distribution.
\end{itemize}

\section{Related Work}

Many techniques have been proposed to train neural networks for calibration. These can be organized into three categories.
One category augments or replaces the primary training loss with a term to explicitly incentivize calibration. These include the AvUC loss \citep{Krishnan2020ImprovingMC}, MMCE loss \citep{Kumar2018TrainableCM} and Focal loss \citep{Mukhoti2020CalibratingDN, lin2018focal}.
The Mean Squared Error loss also compares favourably \citep{hui2021evaluation} to cross-entropy loss. We show that our methods outperform all these calibration-incentivizing training objectives, applied across multiple primary losses.
Label smoothing \citep{muller2020does} has been shown to improve calibration and can be interpreted as a modified primary loss.

A second category of methods are post-hoc calibration methods, which rescale model predictions after training. These methods optimize additional parameters on a held-out validation set \citep{Platt1999ProbabilisticOF, Zadrozny2002TransformingCS, Kull2019BeyondTS, Zadrozny2001ObtainingCP, Naeini2016BinaryCC, Allikivi2019NonparametricBI, Kull2017BetaCA, Naeini2015ObtainingWC, Wenger2020NonParametricCF,Gupta2020CalibrationON}. 
The most popular technique is temperature scaling \citep{Guo2017OnCO}, which maximizes a single temperature parameter on held-out NLL. We examine temperature scaling and propose an improvisation that directly optimizes temperature for a soft calibration objective instead of NLL. Temperature scaling has shown to be ineffective under distribution shift in certain scenarios \citep{Ovadia2019CanYT}. We show that uncertainty estimates of methods which train for calibration generalize better than temperature scaling under distribution shift.

A third category of methods examines model changes such as ensembling multiple predictions \citep{Lakshminarayanan2017SimpleAS,wen2020batchensemble} or priors \citep{dusenberry2020efficient}. 
Similar to previous work \citep{Kumar2018TrainableCM, lin2018focal}, we focus on the choice of loss functions for improving calibration of a single neural network---whether during training or post-hoc---and  
do not compare against ensemble models or Bayesian neural networks.
These techniques are complementary to ours and can be combined with our techniques to further improve performance.

Recent works have investigated issues \citep{nixon2019measuring,Kumar2019VerifiedUC,Roelofs2020MitigatingBI, Gupta2020CalibrationON} with the originally proposed ECE \citep{Guo2017OnCO} and suggested new ones. Debiased CE \citep{Kumar2019VerifiedUC} and mean-sweep CE \citep{Roelofs2020MitigatingBI} have been shown to have lesser bias and more consistency across the number of bins parameter than ECE whereas KS-error \citep{Gupta2020CalibrationON} avoids binning altogether. We report these metrics in the Appendix.

%





\section{Background}
\label{sec:background}
\subsection{The Task and the Model}

Consider a classification task over $K$ classes with a dataset of $N$ samples $D= \langle (\vx_i,y_i) \rangle _{i=1}^N$ drawn from the joint probability distribution $\mathcal D (\mathcal X, \mathcal Y)$ over the input space $\mathcal X$ and label space $\mathcal Y=\{1,2,\ldots K\}$. The task is modelled using a deep neural network with parameters $\vtheta$ whose top layer is interpreted as a softmax layer. The top layer consists of $K$ neurons which produce logits $\vg_{\vtheta}(\vx) = \langle g_{\vtheta}(y|\vx) \rangle_{y \in \mathcal Y}$. 
The predictive probabilities for a given input are:%
$$  \vf_{\vtheta}(\vx) = \langle f_{\vtheta}(y|\vx) \rangle_{y \in \mathcal Y} = \softmax(\vg_{\vtheta}(\vx)).$$
The parameters $\vtheta$ of the neural network are trained to minimize $\mathbb E_{(\vx, y)} \mathcal L (\vf_{\vtheta}(\vx), y)$ where $(\vx, y)$ is sampled from $\mathcal D (\mathcal X, \mathcal Y)$. Here $\mathcal L$ is a trainable loss function which incentivizes the predictive distribution $\vf_{\vtheta}(\vx)$ to fit the label $y$. The model's prediction on datapoint $(\vx, y)$ is denoted by $q_{\vtheta}(\vx) = \arg\max (\vf_{\vtheta}(\vx))$. We denote by $c_{\vtheta}(\vx) = \max (\vf_{\vtheta}(\vx))$ the confidence of this prediction and by boolean quantity $a_{\vtheta}(\vx, y) = \textbf{1}_{q_{\vtheta}(\vx) = y}$ the accuracy of this prediction.


\subsection{Expected Calibration Error (ECE)}

Given a distribution $\hat {\mathcal D} (\mathcal X, \mathcal Y)$ on datapoints, there are two standard notions of Ideal Calibration Error that we refer to as Ideal Binned Expected Calibration Error (of order $p$), $\textrm{IECE}_{\textrm{bin},p}(\hat {\mathcal D}, \vtheta) = \textrm{IECE}_{\textrm{bin},p}$, and Ideal Expected Label-Binned Calibration Error (of order $p$), $\textrm{IECE}_{\textrm{lb},p}(\hat {\mathcal D}, \vtheta) = \textrm{IECE}_{\textrm{lb},p}$. This nomenclature is consistent with that introduced by \citet{Roelofs2020MitigatingBI}. Both these measure, in slightly different ways, the $p^\text{th}$ root of the $p^\text{th}$ moment of the absolute difference between model confidence and the empirical accuracy given that confidence. They are defined as follows:
\begin{align}
\textrm{IECE}_{\textrm{bin},p}(\hat {\mathcal D}, \vtheta) &= \Big (\mathbb E_{c_{\vtheta}(\hat {\vx_0})}\Big [|\mathbb E[a_{\vtheta}(\hat {\vx_1}, \hat {y_1})|c_{\vtheta}(\hat {\vx_1}) = c_{\vtheta}(\hat {\vx_0})] - c_{\vtheta}(\hat {\vx_0})|^p\Big ] \Big )^{1/p} \label{eq:iece-bin}\\
\textrm{IECE}_{\textrm{lb},p}(\hat {\mathcal D}, \vtheta) &= \Big ( \mathbb E_{({\hat {\vx_0}}, \hat {y_0})} \Big [|\mathbb E[a_{\vtheta}(\hat {\vx_1}, \hat {y_1})|c_{\vtheta}(\hat{\vx_1}) = c_{\vtheta}({\hat {\vx_0}})] - c_{\vtheta}({\hat {\vx_0}})|^p \Big ] \Big )^{1/p}. \label{eq:iece-lb}
\end{align}
Note that the critical dependence on $\hat {\mathcal D} (\mathcal X, \mathcal Y)$ is implicit in both definitions since the datapoint $(\hat {\vx_0}, \hat {y_0})$ from the outer expectation and the datapoint $(\hat {\vx_1}, \hat {y_1})$ from the inner expectation are both sampled from $\hat {\mathcal D} (\mathcal X, \mathcal Y)$.


We cannot compute $\textrm{IECE}_{\textrm{bin},p}$ and $\textrm{IECE}_{\textrm{lb},p}$ in practice since the number of datapoints are finite. Instead we consider a dataset $\hat D= \langle (\hat{\vx_i}, \hat {y_i}) \rangle _{i=1}^{\hat N}$ drawn from $\hat {\mathcal D} (\mathcal X, \mathcal Y)$ and partition the confidence interval $[0,1]$ into bins $\mathcal B = \langle B_i \rangle_{i \in \{1,2,\ldots M\}}$, each of which also corresponds to a confidence interval. We will use $c_i$ as a shorthand for $c_{\vtheta}(\hat{\vx_i})$ and $a_i$ as a shorthand for $a_{\vtheta}(\hat{\vx_i}, \hat {y_i})$. We denote by $b_i(\mathcal B,\hat D,\vtheta) = b_i$ the bin to which $c_i$ belongs. We define the size of bin $j$ as $S_j(\mathcal B,\hat D,\vtheta) = S_j$, the average confidence of bin $j$ as $C_j(\mathcal B,\hat D,\vtheta) = C_j$ and the average accuracy of bin $j$ as $A_j(\mathcal B,\hat D,\vtheta) = A_j$. These are expressed as follows:
\begin{align}
S_j(\mathcal B,\hat D,\vtheta) &= |\{i | b_i = j\}| \label{eq:bin-size-hard}\\
C_j(\mathcal B,\hat D,\vtheta) &= \tfrac{1}{S_j} \Sigma_{i | b_i = j} c_i \label{eq:bin-conf-hard}\\
A_j(\mathcal B,\hat D,\vtheta) &= \tfrac{1}{S_j} \Sigma_{i | b_i = j} a_i. \label{eq:bin-acc-hard}
\end{align}
We are now in a position to define the Expected Binned Calibration Error of order $p$ which we denote by $\textrm{ECE}_{bin,p}$ and Expected Label-Binned Calibration Error of order $p$ we denote by $\textrm{ECE}_{lb,p}$. These serve as empirical approximations to the corresponding intractable ideal notions from equations \ref{eq:iece-bin} and \ref{eq:iece-lb}. They are defined as follows:
\begin{align}
\textrm{ECE}_{bin,p}(\mathcal B, \hat D,\vtheta) &= \Big ( \Sigma_{i=1}^M \tfrac{S_j}{\hat N} \cdot |A_j - C_j|^p \Big )^{1/p} \label{eq:ece-bin-hard}\\
\textrm{ECE}_{lb,p}(\mathcal B, \hat D,\vtheta) &= \Big (\tfrac{1}{\hat N}  \Sigma_{i=1}^{\hat N} |A_{b_i} - c_i|^p \Big )^{1/p}. \label{eq:ece-lb-hard}
\end{align}

It follows from Jensen's inequality that $\textrm{ECE}_{lb,p}(\mathcal B, D,\vtheta) \geq \textrm{ECE}_{bin,p}(\mathcal B, D,\vtheta)$
\citep{Roelofs2020MitigatingBI}.

\section{Soft Calibration Objectives}
\label{sec:soft-calibration-objectives}

In this section, we define quantities that can be used to better incentivize calibration during training.

\subsection{Soft-Binned ECE (SB-ECE)}
\label{sec:sb-ece}

The quantities in the definitions of $\textrm{ECE}_{bin,p}$ and $\textrm{ECE}_{lb,p}$ can be written in terms of a formal definition of the bin membership function. Let us denote the bin-membership function for a given binning $\mathcal B = \langle B_i \rangle_{i \in \{1,2,\ldots M\}}$ by $\vu_{\mathcal B}: [0,1] \rightarrow \mathcal U_M$, where $\mathcal U_M = \{\vv \in \mathbb [0,1]^M | \Sigma_jv_j = 1\}$ is the set of possible bin membership vectors over $M$ bins. The membership function for bin $j$ is denoted by $u_{\mathcal B,j}: [0,1] \rightarrow [0,1]$ and is defined by $u_{\mathcal B,j}(c) = {\vu_{\mathcal B}(c)}_j$. The size, average accuracy, and average confidence of bin $j$ from equations \ref{eq:bin-size-hard}, \ref{eq:bin-conf-hard}, and \ref{eq:bin-acc-hard} can now be written in terms of $\vu_{\mathcal B}$ as follows:
\begin{align}
S_j(\mathcal B,\hat D,\vtheta) &= \Sigma_{i=1}^{\hat N} u_{\mathcal B, j}(c_i) \label{eq:bin-size-soft} \\
C_j(\mathcal B,\hat D,\vtheta) &= \tfrac{1}{S_j}  \Sigma_{i=1}^{\hat N} (u_{\mathcal B, j}(c_i) \cdot c_i) \label{eq:bin-conf-soft} \\
A_j(\mathcal B,\hat D,\vtheta) &= \tfrac{1}{S_j} \Sigma_{i=1}^{\hat N} (u_{\mathcal B, j}(c_i) \cdot a_i). \label{eq:bin-acc-soft}
\end{align}
The quantities $\textrm{ECE}_{bin,p}$ and $\textrm{ECE}_{lb,p}$ can further be written in terms of the quantities $S_j$, $C_j$ and $A_j$ using equation \ref{eq:ece-bin-hard} and a modification of equation \ref{eq:ece-lb-hard} (see equation \ref{eq:ece-lb-soft} below). We know that the differentials $\partial \textrm{ECE}_{bin,p} / \partial \vtheta$ and $\partial \textrm{ECE}_{lb,p} / \partial \vtheta$ are non-trainable. The formulation above makes it clear that this is precisely because $\partial u_{\mathcal B, j}/ \partial c$ is zero within bin boundaries and undefined at bin boundaries. Moreover, this observation implies that if we could come up with a trainable soft bin-membership function $\vu^*_{\mathcal B}$ then we could use it in place of the usual hard bin-membership function $\vu_{\mathcal B}$ to obtain a trainable version of $\textrm{ECE}_{bin,p}$ and $\textrm{ECE}_{lb,p}$.

With this motivation, we define the soft bin-membership function that has a well-defined non-zero gradient in $(0,1)$. It is parameterized by the number of bins $M$ and a temperature parameter $T$. We consider equal-width binning for simplicity and so we represent it as $\vu^*_{M,T}$ rather than $\vu^*_{\mathcal B}$. We desire unimodality over confidence: if $\xi_j$ denotes the center of bin $j$ then we want $\partial {u^*_{M,T,j}} / \partial c$ to be positive for $c < \xi_j$ negative for $c > \xi_j$. Similarly, we also desire unimodality over bins: if $c < \xi_i < \xi_j$ or $c > \xi_i > \xi_j$, then we want that ${u^*_{M,T,i}}(c) > {u^*_{M,T,j}}(c)$. Finally, we also want the aforementioned temperature parameter $T$ to control how close the binning is to hard binning (i.e. how steeply membership drops off). This would give us the nice property of hard-binning being a limiting condition of soft-binning. With this motivation, we define the soft bin-membership function as
\begin{align*}
\vu^*_{M,T}(c) &= \softmax({\vg_{M,T}}(c)), \\
\text{where } g_{M,T,i}(c) &= -(c-\xi_i)^2 / T & \forall i \in \{1,2,\ldots M\}.
\end{align*}
%
%
%
Figure \ref{fig-sb:ece} visualizes soft bin-membership.
We can now formulate trainable calibration error measures. We define the Expected Soft-Binned Calibration Error $\textrm{SB-ECE}_{\textrm{bin},p}(M, T, \hat D,\vtheta)$
and Expected Soft-Label-Binned Calibration Error $\textrm{SB-ECE}_{\textrm{lb},p}(M, T, \hat D,\vtheta)$:
%
\begin{SCfigure}
   \centering
   \caption{Visualization of the soft bin membership function which shows that the temperature parameter determines the sharpness of the binning. Soft binning limits to hard binning as temperature tends to zero.}
   \includegraphics[width=0.7\textwidth]{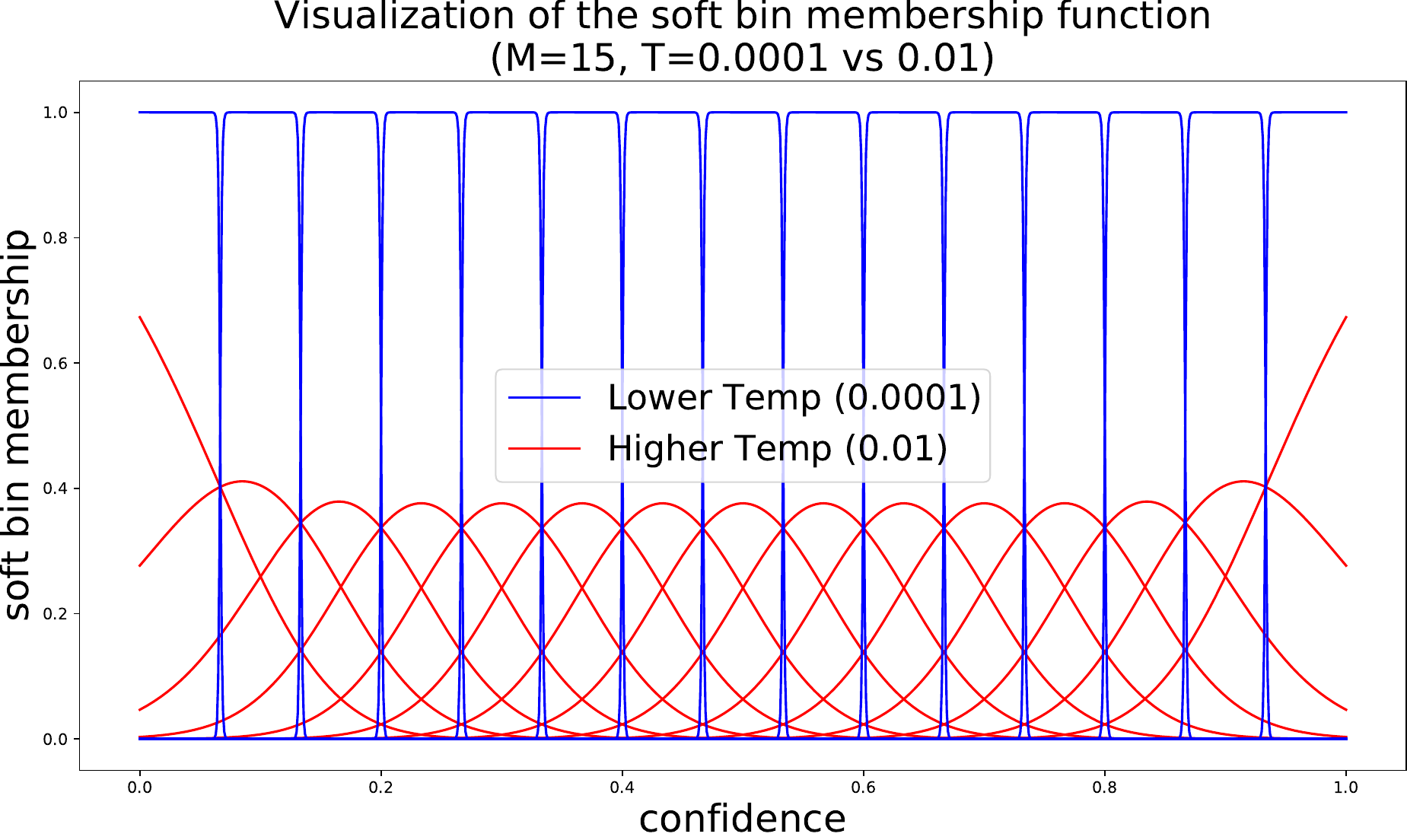}
   \label{fig-sb:ece}
\end{SCfigure}
\begin{align}
\textrm{SB-ECE}_{\textrm{bin},p}(M, T, \hat D,\vtheta) &= \Big ( \Sigma_{i=1}^M (\tfrac{S_j}{\hat N}  |A_j - C_j|^p) \Big )^{1/p}, \label{eq:ece-bin-soft}\\
\textrm{SB-ECE}_{\textrm{lb},p}(M, T, \hat D,\vtheta) &= \Big (\tfrac{1}{\hat N} \Sigma_{i=1}^{\hat N} \Sigma_{j=1}^M (u^*_{\mathcal M, T, j}(c_i) \cdot |A_j - c_i|^p )\Big )^{1/p}. \label{eq:ece-lb-soft}
\end{align}
The quantities $S_j$, $C_j$ and $A_j$ in these expressions are obtained by using the soft bin membership function $\vu^*_{M,T}$ in place of the hard bin membership function $\vu_{\mathcal B}$ in equations \ref{eq:bin-size-soft}, \ref{eq:bin-conf-soft} and \ref{eq:bin-acc-soft} respectively.

We use these trainable calibration error measures as: (1) part of the training loss function and (2) the objective that is minimized to tune the temperature scaling parameter $T_{ts}$ during post-hoc calibration.


\subsection{Soft AvUC (S-AvUC)}

The accuracy versus uncertainty calibration (AvUC) loss \citep{Krishnan2020ImprovingMC} categorizes each prediction that a model with parameters $\vtheta$ makes for a labelled datapoint $d_i = (\vx_i, y_i)$ from dataset $\hat D$ according to two axes: (1) accurate [A] versus inaccurate [I], based on the value of the boolean quantity $a_{\vtheta}(\vx_i, y_i) = a_i$ (2) certain [C] versus uncertain [U], based on whether the entropy $h(f_{\vtheta}(\vx_i)) = h_i$ of the predictive distribution is above or below a threshold $\kappa$. Denote the number of elements from $\hat D$ that fall in each of these 4 categories by $\hat n_\textrm{AC}$, $\hat n_\textrm{AU}$, $\hat n_\textrm{IC}$ and $\hat n_\textrm{IU}$ respectively.
The AvUC loss incentivizes the model to be certain when accurate and uncertain when inaccurate:
\vspace{1ex}
\begin{align}
\textrm{AvUC}(\kappa, \hat D, \vtheta) = \log\Big(1 + \frac{n_\textrm{AU} + n_\textrm{IC}}{n_\textrm{AC} + n_\textrm{IU}}\Big),
\label{eq:avuc}
\end{align}
where the discrete quantities are relaxed to be differentiable:
\begin{equation}
  \begin{split}
    n_\textrm{AU} &= \Sigma_{i|(\vx_i, y_i) \in S_\textrm{AU}} (c_i \tanh h_i) \\
    n_\textrm{AC} &= \Sigma_{i|(\vx_i, y_i) \in S_\textrm{AC}} (c_i (1 - \tanh h_i))
  \end{split}
\quad\quad
  \begin{split}
    n_\textrm{IC} &= \Sigma_{i|(\vx_i, y_i) \in S_\textrm{IC}} ((1 - c_i) (1 - \tanh h_i)) \\
    n_\textrm{IU} &= \Sigma_{i|(\vx_i, y_i) \in S_\textrm{IU}} ((1 - c_i) \tanh h_i).
  \end{split}
\label{eq:avuc-nxx}
\end{equation}
\citet{Krishnan2020ImprovingMC} have showed good calibration results using the AvUC loss in SVI settings. However, in our experiments we found that the addition of the AvUC loss term resulted in poorly calibrated models in non-SVI neural network settings (see Appendix \ref{sec:full-table-of-results}). One reason for this seems to be that minimizing the AvUC loss results in the model being incentivized to be even more confident in its inaccurate and certain predictions (via minimizing $n_\textrm{IC}$) and even less confident in its accurate and uncertain predictions (via minimizing $n_\textrm{AU}$). This conjecture is validated by experimental observations: when we stopped the gradients flowing through the $c_i$ terms in equation \ref{eq:avuc-nxx}, we were able to obtain calibrated models (see Appendix \ref{sec:avuc-gs}). Fixing this incentivization issue in a more principled manner than stopping gradients is desirable. Another desirable improvisation is replacing the hard categorization into the certain/uncertain bins with a soft partitioning scheme. We meet both these objectives by defining a notion of soft uncertainty.

We want a limiting case of the soft uncertainty function to be the hard uncertainty function based on an entropy threshold $\kappa$. This implies that we will continue to have a parameter $\kappa$ despite getting rid of the hard threshold. As before, we desire a temperature parameter $T$ that will determine how close the function is to the hard uncertainty function. The soft uncertainty function $t_{\kappa,T}:[0,1]\rightarrow [0,1]$ takes as input the $[0,1]$-normalized entropy $h^*_i = h_i / \log(K)$ of the predicted posterior where $K$ is the number of classes. We also need $\partial t_{\kappa,T} / \partial {h^*}$ to be positive in $[0,1]$ and would like $t_{\kappa,T}$ to satisfy $\lim_{h^*\rightarrow 0}t_{\kappa,T}(h^*)=0$ and $\lim_{h^*\rightarrow 1}t_{\kappa,T}(h^*)=1$. Finally, it would be good to have the $[0,1]$ identity mapping as a special case of $t_{\kappa,T}$-family for some value of $\kappa$ and $T$. We now define the soft-uncertainty function in the following way so that it meets all stated desiderata:
\vspace{1ex}
\begin{align*}
t_{\kappa,T}(h^*) &= \text{logistic}\left(\frac{1}{T} \log\frac{h^*(1-\kappa)}{(1-h^*)\kappa}\right).
\end{align*}
%
%
\begin{figure}[bt!]
   \centering
    \includegraphics[width=\textwidth]{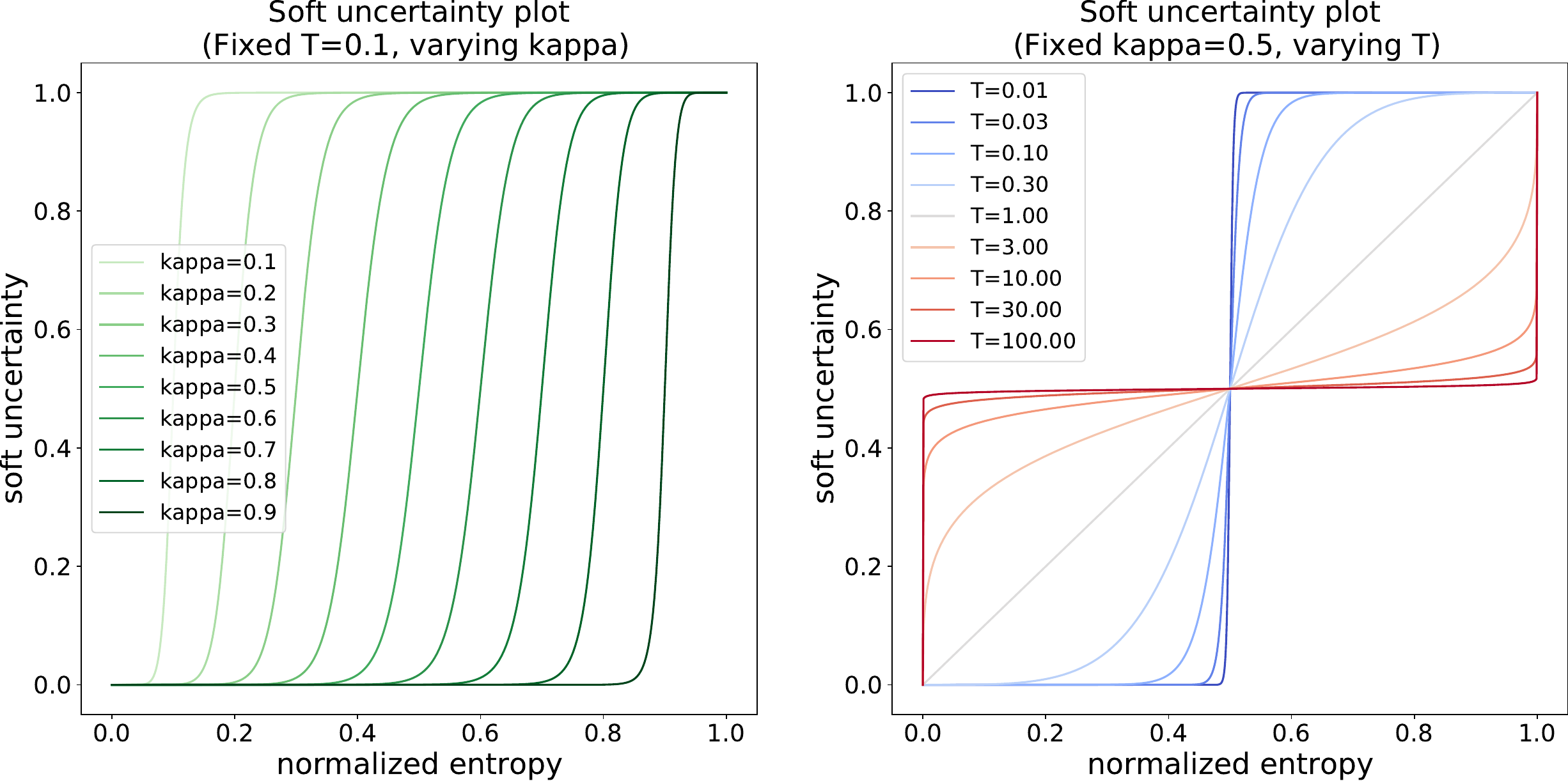}
   \caption{Visualization of the soft uncertainty function $t_{\kappa,T}(h^*)$ which shows that the parameter $\kappa$ captures the soft-threshold whereas the parameter $T$ captures the sharpness of the thresholding.
   }
\end{figure}
Finally, we define Soft AvUC in terms of soft uncertainty by modifying equations \ref{eq:avuc} and \ref{eq:avuc-nxx}. In our experiments, we use Soft AvUC as part of the loss function to obtain calibrated models.
\begin{align}
\textrm{S-AvUC}(\kappa, T, \hat D, \vtheta) = \log\Big(1 + \frac{n'_\textrm{AU} + n'_\textrm{IC}}{n'_\textrm{AC} + n'_\textrm{IU}}\Big),
\label{eq:svuc}
\end{align}
where
\begin{equation}
  \begin{split}
    n'_\textrm{AU} &= \Sigma_{i|a_i=1} (t_{\kappa,T}(h^*_i) \tanh h_i) \\
    n'_\textrm{AC} &= \Sigma_{i|a_i=1} ((1 - t_{\kappa,T}(h^*_i)) (1 - \tanh h_i))
  \end{split}
\quad\quad
  \begin{split}
    n'_\textrm{IC} &= \Sigma_{i|a_i=0} ((1 - t_{\kappa,T}(h^*_i)) (1 - \tanh h_i)) \\
    n'_\textrm{IU} &= \Sigma_{i|a_i=0} (t_{\kappa,T}(h^*_i) \tanh h_i).
  \end{split}
\label{eq:savuc-nxx}
\vspace{-1ex}
\end{equation}
%

\section{Results}

We compare our Soft Calibration Objectives to recently proposed calibration-incentivizing training objectives MMCE, focal loss, and AvUC on the CIFAR-10, CIFAR-100, and ImageNet datasets.
We evaluate the full cross-product of primary and secondary losses: the options for primary loss are cross-entropy (NLL), focal or mean squared error (MSE) loss; and the options for secondary loss are MMCE, AvUC, SB-ECE or S-AvUC. Results for the MSE primary loss and the AvUC secondary loss are in Appendix \ref{sec:full-table-of-results}. 
Our experiments build on the Uncertainty Baselines and Robustness Metrics libraries \citep{nado2021uncertainty,djolonga2020robustness}.

\begin{figure}[bt!]
   \centering
    \includegraphics[width=\textwidth]{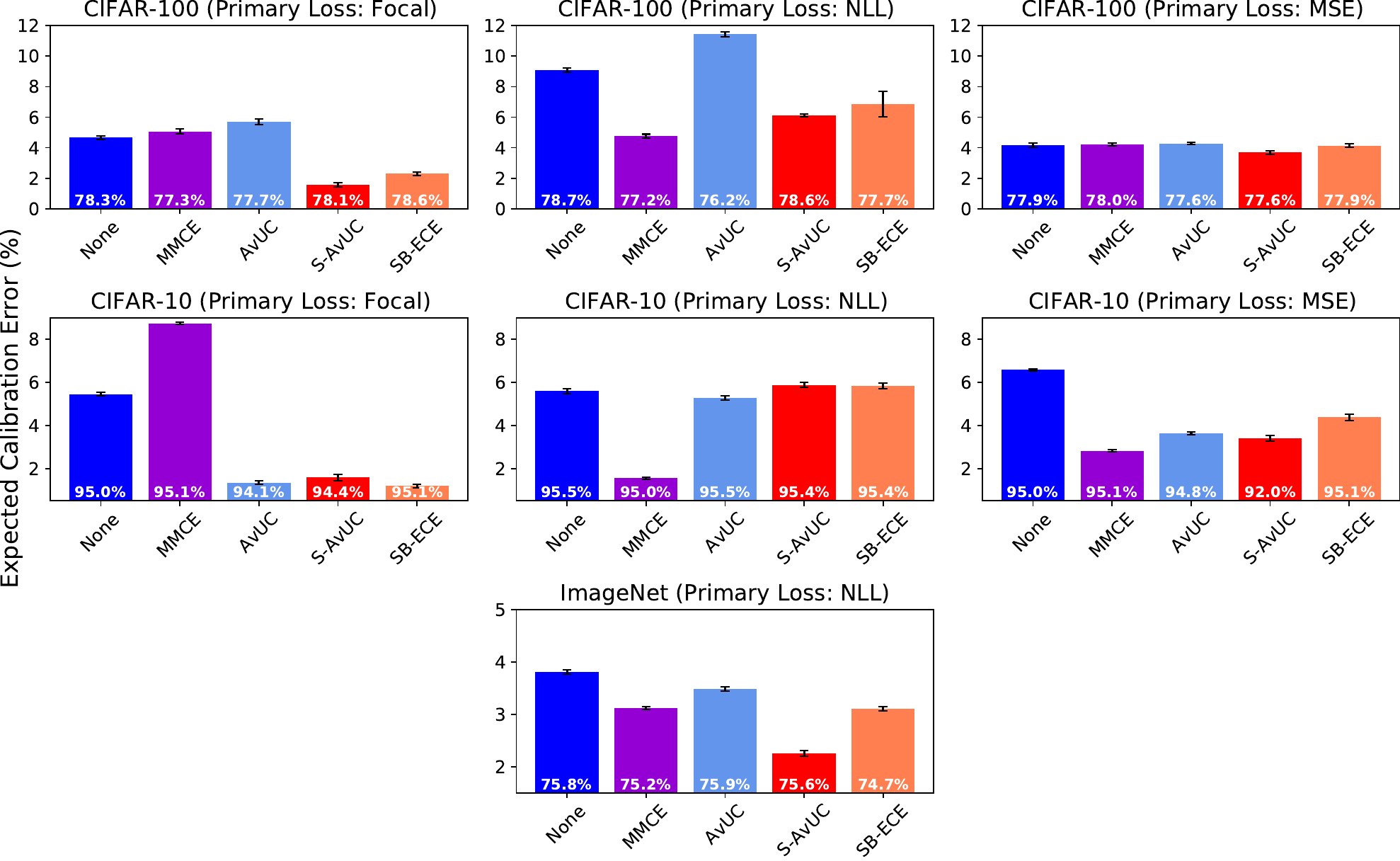}
  \caption{Soft Calibration Objectives (\red{S-AvUC}, \orange{SB-ECE}), when used as secondary losses with a primary loss, achieve lower ECE (equal-mass binning, $\ell_2$ norm) than the corresponding primary loss for CIFAR-10, CIFAR-100, and ImageNet. These statistically significant wins come at the cost of less than 1\% accuracy (reported at bottom of the bar). Values reported are mean over 10 runs, and the error bars indicate $\pm1$ standard error of the mean (SEM). For each dataset (each row) the best (across primary losses) ECE obtained using the \red{S-AvUC} and \orange{SB-ECE} secondary losses is lower than the best ECE obtained using existing techniques.}
  \label{fig:result-overview}
\end{figure}

\subsection{Soft Calibration Objectives for End-to-End Training}
\label{sec:caltrain-e2e}

Our results demonstrate that training losses which include Soft Calibration Objectives obtain state-of-the-art single-model ECE on the test set in exchange for less than 1\% reduction in accuracy for all three datasets that we experiment with. In fact, our methods (especially S-AvUC) without post-hoc temperature scaling are better than or as good as other methods with or without post-hoc temperature scaling on all three datasets. 

The primary losses we work with for CIFAR-10/100 are the cross-entropy (NLL) loss, the focal loss and mean squared error (MSE) loss. Focal loss \citep{Mukhoti2020CalibratingDN} and MSE loss \citep{hui2021evaluation} have recently shown to outperform the NLL loss in certain settings. The cross-entropy loss outperforms the other two losses on Imagenet, and is thus our sole focus for this dataset.

The primary loss even by itself (especially NLL) can overfit to the train ECE \citep{Mukhoti2020CalibratingDN}, without help from the soft calibration losses. Even in such settings, we show that soft calibration losses yield reduction of test ECE using a technique we call `interleaved training' (Appendix \ref{sec:interleaving}).

We use the Wide-Resnet-28-10 architecture \citep{zagoruyko2017wide} trained for 200 epochs on CIFAR-100 and CIFAR-10. For Imagenet, we use the Resnet-50 \citep{he2015deep} architecture training for 90 epochs. All our experiments use the SGD with momentum optimizer with momentum fixed to 0.9 and learning rate fixed to 0.1. The loss function we use in our experiments is $\textrm{PL} + \beta \cdot \textrm{SL} + \lambda \cdot \textrm{L2}$ where PL and SL denote the primary and secondary losses respectively and L2 denotes the weight normalization term with $\ell_2$ norm. We tune the $\beta$ and $\lambda$ parameters along with the parameters $\kappa$ and $T$ relevant to the secondary losses $\textrm{SB-ECE}_{lb,p}(M, T, \hat D,\vtheta)$ and $\textrm{S-AvUC}(\kappa, T, \hat D, \vtheta)$. We tune these hyperparameters sequentially. We fix the learning rate schedule and the number of bins $M$ to keep the search space manageable. Appendix \ref{sec:hyperparam-tuning} has more details of our hyperparameter search.

\begin{SCfigure}
   \centering
   \caption{Accuracy vs Confidence plots for various methods on CIFAR-100. \blue{NLL} is significantly overconfident and \purple{NLL + MMCE} is somewhat overconfident. While \green{Focal loss} is underconfident, augmenting it with \red{Soft Calibration Objectives} fixes this issue, resulting in curves closest to the ideal.}
   \includegraphics[width=0.7\textwidth]{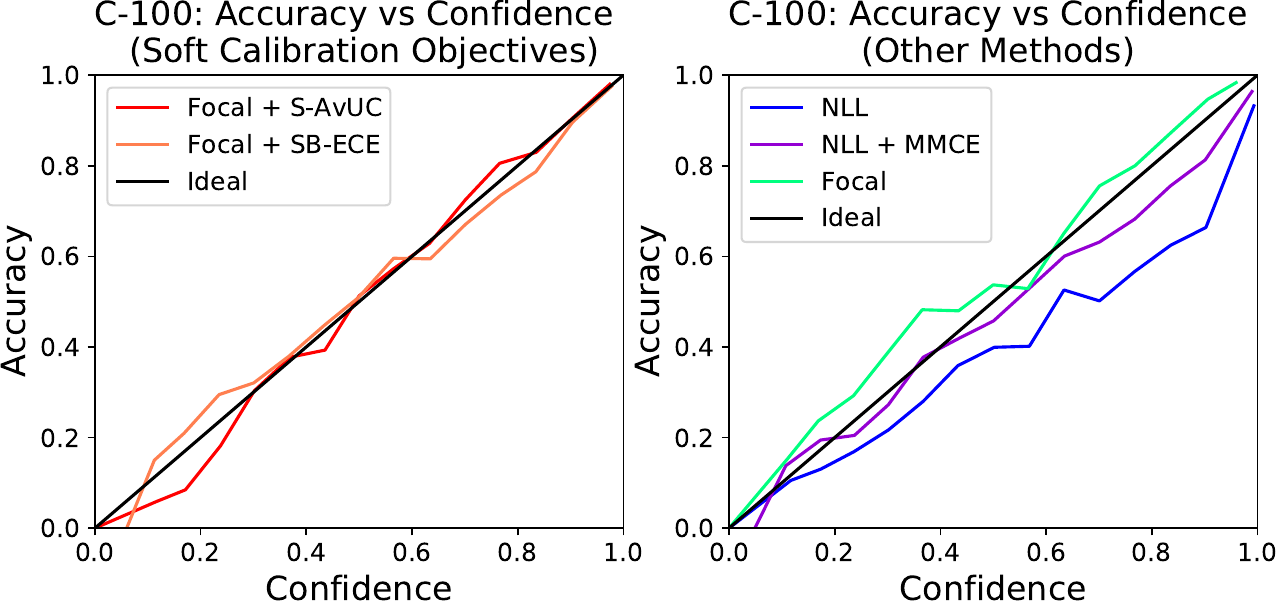}
\label{fig:acc-vs-conf}
\end{SCfigure}

In Table \ref{tab:summary_results}, we report the runs with the best ECE values that also came within 1\% of the primary loss run with the highest accuracy. Figure \ref{fig:acc-vs-conf} is the accuracy-confidence plot corresponding to Table \ref{cifar100-table}. 
In Figure \ref{fig:result-overview}, we visualize the ECE on the test set for all combinations of primary loss, secondary loss and dataset. The best ECE for each dataset is attained using Soft Calibration Objectives as secondary losses. 
%
%
%
More such figures are in Appendix \ref{sec:reliability-plots} and the complete table can be found in Appendix \ref{sec:full-table-of-results}. 

\begin{table}[h]
\caption{ We report average accuracy (with standard error across 10 trials), ECE, and ECE obtained after post-hoc temperature scaling (TS) for models trained with different objectives on the CIFAR-10, CIFAR-100, and ImageNet datasets. ECE is computed with the $\ell_2$ norm and equal-mass binning. We find Soft Calibration Objectives (SB-ECE, S-AvUC) result in better or equivalent ECEs compared to previous methods with or without TS.  We also find that TS does not always improve ECE. The best ECE value is highlighted for each dataset. The best ECE with TS value is highlighted if it improves over the best value from the ECE column.}
\begin{subtable}[c]{0.5\textwidth}
  \subcaption{CIFAR-100}
\label{cifar100-table}
  \centering
  \resizebox{\textwidth}{!}{
  \begin{tabular}{||c||c||c|c||}
    \hline
    Loss Fn. & Accuracy & ECE & \makecell {ECE with  TS} \\
    \hline
    NLL & 78.7$\pm$0.122 & 9.10$\pm$0.139 & 5.36$\pm$0.091 \\
    \hline
    \makecell {NLL +  MMCE} & 77.2$\pm$0.072 & 4.77$\pm$0.121 & 4.06$\pm$0.138 \\
    \hline
    Focal & 78.3$\pm$0.086 & 4.66$\pm$0.130 & 6.47$\pm$0.140 \\
    \hline
    \makecell {Focal +  SB-ECE} & 78.6$\pm$0.062 & 2.30$\pm$0.105 & 5.16$\pm$0.108 \\
    \hline
    \makecell {Focal +  S-AvUC} & 78.1$\pm$0.084 & \textbf{1.57}$\pm$0.122 & 4.15$\pm$0.090 \\
    \hline
  \end{tabular}
  }
\end{subtable} \hfill
\begin{subtable}[c]{0.5\textwidth}
\centering
  \subcaption{CIFAR-10}
\label{cifar10-table}
  \resizebox{\textwidth}{!}{
  \begin{tabular}{||c||c||c|c||}
    \hline
    Loss Fn. & Accuracy & ECE & \makecell {ECE with  TS} \\
    \hline
    NLL & 95.5$\pm$0.040 & 5.59$\pm$0.119 & 1.95$\pm$0.127 \\
    \hline
    \makecell {NLL +  MMCE} & 95.0$\pm$0.031 & 1.55$\pm$0.053 & \textbf{1.09}$\pm$ 0.098 \\
    \hline
    Focal & 95.0$\pm$ 0.085 & 5.45$\pm$0.079 & 2.69$\pm$0.190 \\
    \hline
   \makecell {Focal +  SB-ECE} & 95.1$\pm$ 0.056 & \textbf{1.19}$\pm$ 0.088 & 2.08$\pm$ 0.143 \\
    \hline
    \makecell {Focal + S-AvUC} & 94.4$\pm$ 0.145 & 1.58$\pm$ 0.146 & 1.34$\pm$ 0.172 \\
    \hline
  \end{tabular}
  }
\end{subtable} \\
\begin{subtable}[c]{0.5\textwidth}
\centering
  \caption{ImageNet}
    \resizebox{\textwidth}{!}{
    \begin{tabular}{||c||c||c|c||}
    \hline
    Loss Fn. & Accuracy & ECE & ECE with TS \\
    \hline
    NLL & 75.8$\pm$ 0.036 & 3.81$\pm$ 0.043 & 2.17$\pm$ 0.045 \\
    \hline
    NLL + MMCE & 75.2$\pm$ 0.048 & 3.12$\pm$ 0.025 & 2.18$\pm$ 0.029 \\
    \hline
    NLL + SB-ECE & 74.7 $\pm$ 0.028 & 3.11 $\pm$ 0.039 & \textbf{1.92}$\pm$ 0.024 \\
    \hline
    NLL + S-AvUC & 75.6 $\pm$ 0.053 & \textbf{2.26} $\pm$ 0.055 & 2.02 $\pm$ 0.041 \\
    \hline
  \end{tabular}}

    \label{imagenet-table}
\end{subtable}
\label{tab:summary_results}
\end{table}


\subsection{Soft Calibration Objectives for Post-Hoc Calibration}

Standard temperature scaling (TS) uses a cross-entropy objective to optimize the temperature.  However, using our differentiable soft binning calibration objective (SB-ECE), we can optimize the temperature using a loss function designed to directly minimize calibration error.
In Figure \ref{fig:ts-comparison} (and Figure \ref{fig:ts-comparison-appendix} in Appendix \ref{sec:sco-post-hoc-ts}), we compare temperature scaling with a soft binning calibration objective (TS-SB-ECE) to standard temperature scaling with a cross-entropy objective (TS) on out-of-distribution shifts of increasing magnitude on both CIFAR-10 and CIFAR-100. The distribution shifts come from either the CIFAR-10-C or CIFAR-100-C datasets. Whereas Figure \ref{fig:ts-comparison} focuses on cross-entropy loss, Figure \ref{fig:ts-comparison-appendix} also has the plots for other primary losses. Table \ref{tab:all-test-results} contains comparisons between the two methods based on test ECE for all combinations of dataset, primary loss and secondary loss. We find that TS-SB-ECE outperforms TS under shift in most cases, and the performance increase is similar across shifts of varying magnitude. Note that temperature scaling (TS) does not always improve ECE, especially when the training loss is different from NLL. In such cases TS-SB-ECE still outperforms TS but may or may not result in ECE improvement.

\subsection{Training for Calibration Under Distribution Shift}
\label{sec:caltrain-ood}

\begin{figure}[bt!]
   \centering
    \includegraphics[width=\textwidth]{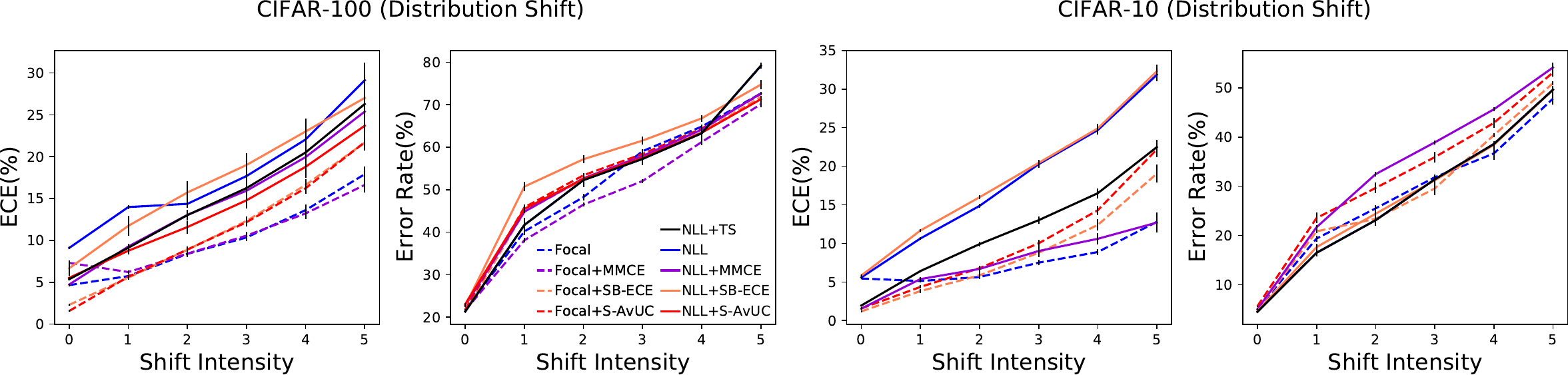}
  \caption{For the CIFAR-10/100 datasets, methods which train for calibration outperform the popular methods - \blue{NLL} and NLL + TS - under distribution shift. \blue{Focal} primary loss and \purple{MMCE} secondary loss result in the lowest ECEs under shift. We note that these methods start off with worse ECEs than our \orange{SB-ECE} and \red{S-AvUC} methods on the test set but end up with better ECE under increasing levels of skew. The OOD datasets that we have used here are CIFAR-10/100-C with skew levels 1-5.}
  \label{fig:cifar-ood}
\end{figure}

\begin{figure}[bt!]
   \centering
    \includegraphics[width=\textwidth]{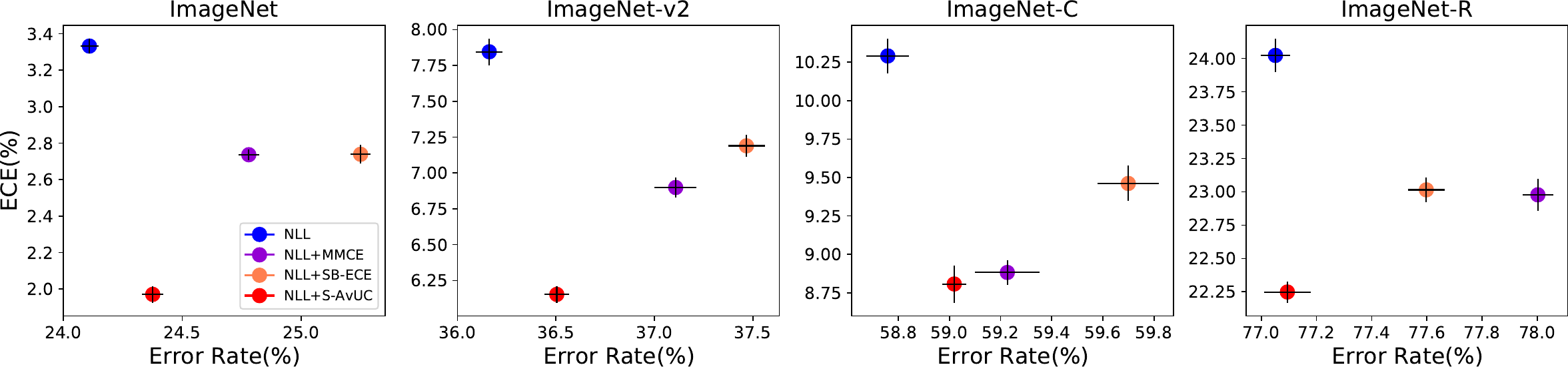}
  \caption{A comparison of methods for ImageNet (primary loss: cross-entropy) shows that the \red{S-AvUC} secondary loss yields lowest ECEs under dataset shift. Error bars are $\pm1$ SEM over 10 runs.}
  \label{fig:imagenet-ood}
\end{figure}

In previous sections we have shown that training for calibration outperforms the popular cross-entropy loss coupled with post-hoc TS on the in-distribution test set. We find that methods which train for calibration (not always our proposed methods) also outperform the cross-entropy loss with TS under dataset shift. Prior work has shown that temperature scaling can perform poorly under distribution shift \citep{Ovadia2019CanYT}, and our experiments reproduce this issue. Moreover, we show that training for calibration makes progress towards fixing this problem. However, different methods perform best under distribution shift on different datasets. Whereas S-AvUC does well on ImageNet OOD (see figure \ref{fig:imagenet-ood}), Focal loss does better than our methods on CIFAR-10-C and CIFAR-100-C (see figure \ref{fig:cifar-ood}). We cannot prescribe one method in particular under distribution shift given these results but we have shown a crucial benefit of using methods to train for calibration as a whole.


\section{Conclusions}
\label{sec:conclusions}

We proposed Soft Calibration Objectives motivated by the goal of directly training models for calibration. These objectives - SB-ECE and S-AvUC - are softened versions of the usual ECE measure and the recently-proposed AvUC loss, respectively. They are easy-to-implement augmentations to the popular cross-entropy loss. We performed a thorough comparison of existing methods of training for calibration. Our experiments show that methods based on soft-calibration objectives can be used to obtain the best ECE among such methods in exchange for less than 1\% drop in accuracy. We note that a model being better calibrated overall does not necessarily mean that it is better calibrated for every group and hence the fairness of our methods as well as related methods must be studied. However, our methods of training-for-calibration can be adapted to encourage fairness by applying the methods separately to each protected group. 

Even when one does not wish to incorporate secondary losses to train for calibration,
we showed that post-hoc temperature scaling works better when tuned using the SB-ECE objective instead of the standard cross-entropy loss. Practitioners can easily replace the cross-entropy loss with our SB-ECE loss when performing post-hoc temperature scaling. 

Finally, we demonstrated that the uncertainty estimates of methods which train for calibration generalize better under dataset shift as compared to post-hoc calibration, which is a fundamental motivation for transitioning to training for calibration.

\section*{Acknowledgements}
The authors thank Brennan McConnell and Mohammad Khajah
who conducted initial explorations of soft binning calibration loss. The authors also thank Zack Nado, D. Sculley and Jeremiah Liu for help with implementation and suggestions for the writeup.

\clearpage
\newpage
\bibliographystyle{plainnat}
\bibliography{references}

\clearpage
\newpage

\appendix

\section{Complete Table of Results}
\label{sec:full-table-of-results}

We perform thorough experimentation for the full cross-product of primary and secondary losses for each dataset. We tune several hyperparameters (section \ref{sec:hyperparam-tuning}) for each of these settings. For the CIFAR-100 and CIFAR-10 datasets we consider three primary loss functions: cross-entropy, MSE, and Focal. For ImageNet, we consider only the cross-entropy primary loss since the other two did not match its performance in our experiments. We consider five secondary losses: MMCE \citep{Kumar2018TrainableCM}, AvUC \citep{Krishnan2020ImprovingMC}, AvUC-GS (section \ref{sec:avuc-gs}), SB-ECE (section 4.1), and S-AvUC (section 4.2). We also consider the setting with no secondary loss. This leads to 18 configurations ($3 \times 6$) for CIFAR-100, 18 configurations ($3 \times 6$) for CIFAR-10 and 6 configurations ($1 \times 6$) for ImageNet for a total of 42 hyperparameter tunings. The results for the ``best'' (section \ref{sec:hyperparam-tuning}) hyperparameter configuration for each of these settings are listed in Table \ref{tab:all-test-results}.

In order to measure calibration of a model we consider (1) the ECE of the trained model, (2) the ECE with standard post-hoc temperature scaling, and (3) the ECE with post-hoc temperature scaling for the SB-ECE objective. The ECE measured here is $\ell_2$ norm equal-mass ECE with 15 bins. We showed in Figure \ref{fig:result-overview-cohen-d} that the S-AvUC loss performs best aggregated across settings of datasets and primary losses. However, there isn't one secondary loss that is always the best for every setting. Even so, we do see that for all three datasets, the best ECE values for the trained model result from Soft Calibration Objectives used as secondary losses. These wins come at the cost of less than 1\% accuracy. Soft Calibration Objectives are also either the best or equivalent to other methods for all datasets when we consider the numbers after temperature scaling. Note that post-hoc temperature scaling doesn't always help and should be used only if applying it results in better calibration than the trained model on some held out dataset.

\begin{table}
\centering
\caption{Results for the full cross-product of experimental settings between primary and secondary losses for CIFAR-100, CIFAR-10 and ImageNet. We report average accuracy, ECE and ECE obtained after post-hoc temperature scaling (TS, TS-SB-ECE) for each of the 42 configurations after hyperparameter tuning. These metrics corresponding to the best hyperparameter configuration for that setting. ECE is computed with the $\ell_2$ norm and equal-mass binning. Each cell reports average $\pm$ SEM across 10 runs. The best ECE numbers for each dataset in each of the three ECE columns are highlighted. We see that the best ECE values for all three datasets are obtained using Soft Calibration Objectives as secondary losses.}
\label{tab:all-test-results}
\vspace{0.5em}
\resizebox{\textwidth}{!}{
\begin{tabular}{||l|l|l||l||l|l|l||}
\hline
         &     &    &           Accuracy &                             ECE &                          ECE+TS &                      ECE+TS-SB-ECE \\
\hline Dataset & Primary & Secondary &                    &                                 &                                 &                                 \\
\hline
CIFAR-100 & NLL & <none> &  78.7 $\pm$ 0.122 &               9.10 $\pm$ 0.139 &               5.36 $\pm$ 0.091 &               4.69 $\pm$ 0.069 \\
         &     & MMCE &  77.2 $\pm$ 0.072 &               4.77 $\pm$ 0.121 &               4.06 $\pm$ 0.138 &               4.11 $\pm$ 0.173 \\
         &     &AvUC&  76.2 $\pm$ 0.146 &               11.4 $\pm$ 0.175 &               4.40 $\pm$ 0.190 &               7.64 $\pm$ 0.187 \\
         &     &AvUC-GS&  78.3 $\pm$ 0.126 &               6.21 $\pm$ 0.084 &               8.36 $\pm$ 0.166 &               5.40 $\pm$ 0.241 \\
         &     & S-AvUC &  78.6 $\pm$ 0.079 &               6.10 $\pm$ 0.095 &               9.13 $\pm$ 0.085 &               5.88 $\pm$ 0.243 \\
         &     & SB-ECE &  77.7 $\pm$ 0.167 &               6.86 $\pm$ 0.839 &  \textbf{3.18} $\pm$ 0.093 &               3.51 $\pm$ 0.314 \\
         \hdashline & Focal & <none> &  78.3 $\pm$ 0.086 &               4.66 $\pm$ 0.130 &               6.47 $\pm$ 0.140 &               4.84 $\pm$ 0.115 \\
         &     & MMCE &  77.3 $\pm$ 0.104 &               5.07 $\pm$ 0.147 &               3.58 $\pm$ 0.098 &  \textbf{2.82} $\pm$ 0.110 \\
         &     &AvUC&  77.7 $\pm$ 0.145 &               5.68 $\pm$ 0.181 &               5.87 $\pm$ 0.182 &               3.93 $\pm$ 0.209 \\
         &     &AvUC-GS&  78.1 $\pm$ 0.056 &               3.38 $\pm$ 0.109 &               4.89 $\pm$ 0.118 &               2.91 $\pm$ 0.122 \\
         &     & S-AvUC &  78.1 $\pm$ 0.084 &  \textbf{1.57} $\pm$ 0.122 &               4.15 $\pm$ 0.090 &               2.89 $\pm$ 0.077 \\
         &     & SB-ECE &  78.6 $\pm$ 0.062 &               2.30 $\pm$ 0.105 &               5.16 $\pm$ 0.108 &               4.10 $\pm$ 0.095 \\
         \hdashline & MSE & <none> &  77.9 $\pm$ 0.089 &               4.17 $\pm$ 0.122 &               5.22 $\pm$ 0.126 &               4.79 $\pm$ 0.281 \\
         &     & MMCE &  78.0 $\pm$ 0.071 &               4.22 $\pm$ 0.085 &               5.06 $\pm$ 0.108 &               4.90 $\pm$ 0.111 \\
         &     &AvUC&  77.6 $\pm$ 0.134 &               4.27 $\pm$ 0.067 &               5.29 $\pm$ 0.125 &               4.42 $\pm$ 0.224 \\
         &     &AvUC-GS&  77.8 $\pm$ 0.041 &               4.31 $\pm$ 0.114 &               5.03 $\pm$ 0.091 &               4.15 $\pm$ 0.135 \\
         &     & S-AvUC &  77.6 $\pm$ 0.104 &               3.69 $\pm$ 0.122 &               4.60 $\pm$ 0.144 &               4.26 $\pm$ 0.322 \\
         &     & SB-ECE &  77.9 $\pm$ 0.071 &               4.14 $\pm$ 0.100 &               5.00 $\pm$ 0.092 &               4.27 $\pm$ 0.184 \\
\hline CIFAR-10 & NLL & <none> &  95.5 $\pm$ 0.040 &               5.59 $\pm$ 0.119 &               1.95 $\pm$ 0.127 &               1.16 $\pm$ 0.106 \\
         &     & MMCE &  95.0 $\pm$ 0.031 &               1.55 $\pm$ 0.053 &  \textbf{1.09} $\pm$ 0.098 &               1.45 $\pm$ 0.114 \\
         &     &AvUC&  95.5 $\pm$ 0.053 &               5.27 $\pm$ 0.101 &               2.39 $\pm$ 0.110 &               1.30 $\pm$ 0.106 \\
         &     &AvUC-GS&  95.6 $\pm$ 0.036 &               4.94 $\pm$ 0.109 &               2.07 $\pm$ 0.129 &               1.31 $\pm$ 0.072 \\
         &     & S-AvUC &  95.4 $\pm$ 0.027 &               5.87 $\pm$ 0.113 &               2.24 $\pm$ 0.130 &               1.20 $\pm$ 0.115 \\
         &     & SB-ECE &  95.4 $\pm$ 0.043 &               5.84 $\pm$ 0.121 &               2.28 $\pm$ 0.091 &               1.27 $\pm$ 0.139 \\
         \hdashline & Focal & <none> &  95.0 $\pm$ 0.085 &               5.45 $\pm$ 0.079 &               2.69 $\pm$ 0.190 &               1.77 $\pm$ 0.115 \\
         &     & MMCE &  95.1 $\pm$ 0.068 &               8.74 $\pm$ 0.059 &               3.13 $\pm$ 0.110 &               2.16 $\pm$ 0.210 \\
         &     &AvUC&  94.1 $\pm$ 0.128 &               1.34 $\pm$ 0.084 &               2.53 $\pm$ 0.124 &               1.12 $\pm$ 0.133 \\
         &     &AvUC-GS&  95.2 $\pm$ 0.063 &               1.39 $\pm$ 0.081 &               2.30 $\pm$ 0.125 &               1.46 $\pm$ 0.108 \\
         &     & S-AvUC &  94.4 $\pm$ 0.145 &               1.58 $\pm$ 0.146 &               1.34 $\pm$ 0.172 &  \textbf{1.05} $\pm$ 0.197 \\
         &     & SB-ECE &  95.1 $\pm$ 0.056 &  \textbf{1.19} $\pm$ 0.088 &               2.08 $\pm$ 0.143 &               1.38 $\pm$ 0.187 \\
         \hdashline & MSE & <none> &  95.0 $\pm$ 0.041 &               6.58 $\pm$ 0.034 &               5.33 $\pm$ 0.122 &               4.43 $\pm$ 0.106 \\
         &     & MMCE &  95.1 $\pm$ 0.034 &               2.82 $\pm$ 0.046 &               3.23 $\pm$ 0.137 &               2.34 $\pm$ 0.141 \\
         &     &AvUC&  94.8 $\pm$ 0.050 &               3.64 $\pm$ 0.066 &               4.72 $\pm$ 0.139 &               4.01 $\pm$ 0.178 \\
         &     &AvUC-GS&  94.8 $\pm$ 0.031 &               5.44 $\pm$ 0.146 &               5.56 $\pm$ 0.151 &               5.20 $\pm$ 0.173 \\
         &     & S-AvUC &  92.0 $\pm$ 0.117 &               3.40 $\pm$ 0.126 &               1.87 $\pm$ 0.080 &               1.50 $\pm$ 0.134 \\
         &     & SB-ECE &  95.1 $\pm$ 0.056 &               4.37 $\pm$ 0.143 &               5.08 $\pm$ 0.159 &               4.26 $\pm$ 0.136 \\
\hline ImageNet & NLL & <none> &  75.8 $\pm$ 0.036 &               3.81 $\pm$ 0.043 &               2.17 $\pm$ 0.045 &               3.32 $\pm$ 0.043 \\
         &     & MMCE &  75.2 $\pm$ 0.048 &               3.12 $\pm$ 0.025 &               2.18 $\pm$ 0.029 &               2.68 $\pm$ 0.022 \\
         &     &AvUC&  75.9 $\pm$ 0.035 &               3.48 $\pm$ 0.041 &               3.37 $\pm$ 0.037 &               3.18 $\pm$ 0.036 \\
         &     &AvUC-GS&  75.8 $\pm$ 0.035 &               3.84 $\pm$ 0.030 &               3.15 $\pm$ 0.028 &               3.44 $\pm$ 0.034 \\
         &     & S-AvUC &  75.6 $\pm$ 0.053 &  \textbf{2.26} $\pm$ 0.055 &               2.02 $\pm$ 0.041 &  \textbf{1.92} $\pm$ 0.046 \\
         &     & SB-ECE &  74.7 $\pm$ 0.028 &               3.11 $\pm$ 0.039 &  \textbf{1.92} $\pm$ 0.024 &               2.62 $\pm$ 0.039 \\
\hline
\end{tabular}}
\end{table}

\section{Train Set Memorization and Interleaved Training}
\label{sec:interleaving}

The NLL primary loss has recently shown to heavily overfit the train ECE \citep{Mukhoti2020CalibratingDN} in some settings. In these cases, it essentially memorizes the train set, achieving near-perfect accuracy and calibration on it without help from any calibration-incentivizing losses. This raises a question about the effectiveness of using soft calibration during training for reducing test ECE in such settings.

We saw this happen on 2 of our 7 dataset + primary loss settings (CIFAR-100/10 datasets with the NLL primary loss; see table \ref{interleaving-table}). As expected, using Soft Calibration Objectives as secondary training losses did not help reduce train ECE here. To fix this issue, we modified the training procedure. The train set was split into two - the `majority' train set and the `held-out' train set. Each epoch was also correspondingly split into two - the first part optimized NLL on the majority train set and the second part optimized Soft Calibration Objectives on the held-out train set. This way we avoided incentivizing something that was already overfit. The second part of each epoch incentivized the predicted distribution to have higher entropy. We call this `interleaved training'.

We observed that Soft Calibration Objectives with interleaved training helped reduce test ECE relative to baseline on CIFAR-100 + NLL, as can be seen in table \ref{tab:all-test-results}. For a fixed primary loss, the recommendation to practitioners is to either (1) have a calibration dataset and use interleaving if it yields better ECE on it or (2) in absence of a calibration dataset, use interleaving if train ECE is suspiciously low (e.g. less than 1\%). Another approach is to replace the primary loss which overfits (e.g. NLL) with one that doesn't (e.g. Focal) and then use Soft Calibration Objectives for further gains.

\section{Soft Calibration Objectives for Post-Hoc Temperature Scaling}
\label{sec:sco-post-hoc-ts}

\begin{figure}[bt!]
\centering
    \begin{subfigure}{\textwidth}
        \centering
        \includegraphics[width=\textwidth]{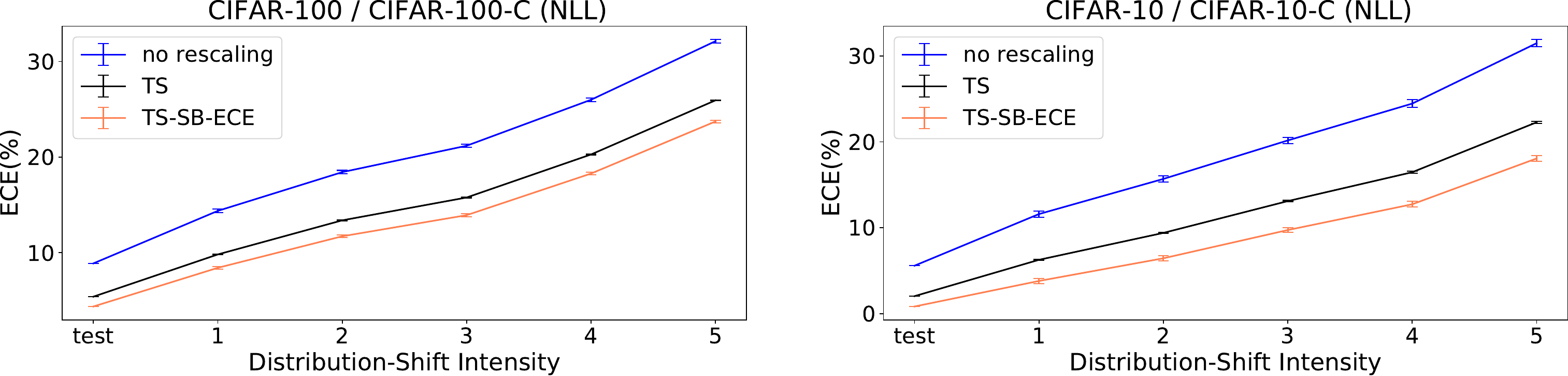}
        \subcaption{NLL}
    \end{subfigure} \\
    \medskip
    \begin{subfigure}{\textwidth}
        \centering
        \includegraphics[width=\textwidth]{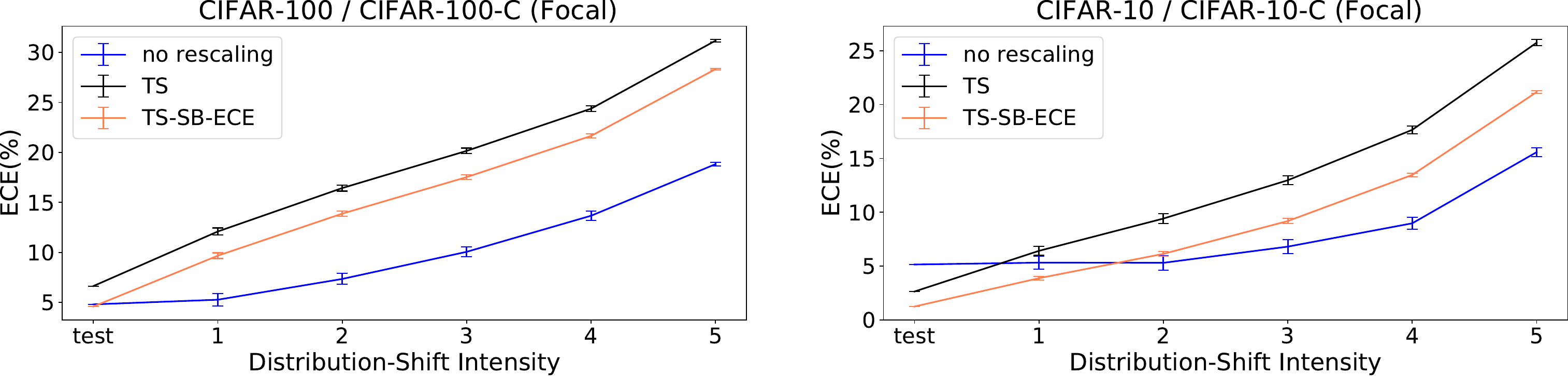}
        \subcaption{Focal}
    \end{subfigure} \\
    \medskip
    \begin{subfigure}{\textwidth}
        \centering
        \includegraphics[width=\textwidth]{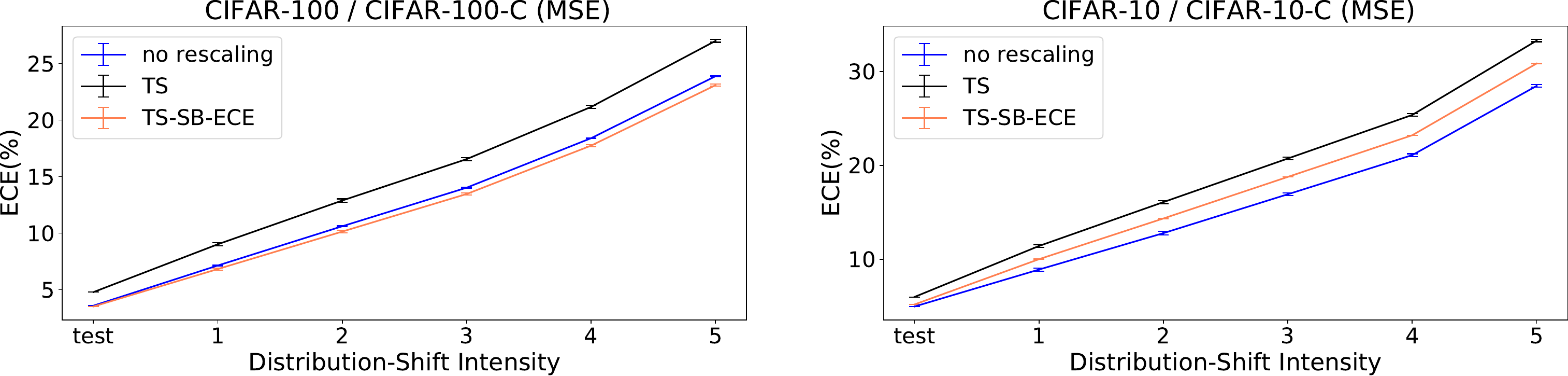}
        \subcaption{MSE}
    \end{subfigure}
    \caption{Post-hoc temperature scaling with the soft calibration error objective (\orange{TS-SB-ECE}) outperforms standard post-hoc temperature scaling (TS), particularly under distribution shift.
  This result holds across datasets (left and right panels),
  distribution shift intensities 
  (along abscissa) and training objectives (rows). The ECE value (equal-mass binning, $\ell_2$ norm) shown is the mean ECE across all corruption types in CIFAR-10-C and CIFAR-100-C. Error bars are
  $\pm1$ standard error of mean (SEM), corrected for intrinsic variability due to type of corruption \citep{MassonLoftus}}.
    \label{fig:ts-comparison-appendix}
\end{figure}

We have seen in Figure \ref{fig:ts-comparison} that temperature scaling for the SB-ECE objective (TS-SB-ECE) outperforms standard temperature scaling (TS) both in- and out-of-distribution for models trained with the popular NLL loss on the CIFAR-100 and CIFAR-10 datasets. We see in figure \ref{fig:ts-comparison-appendix} that this also holds true for the Focal and MSE primary losses. Whereas temperature scaling does not always help to improve calibration, particularly out-of-distribution, we do see that TS-SB-ECE outperforms standard TS in most cases. We plot comparisons for only 6 of our 42 settings in figure \ref{fig:ts-comparison-appendix} for the sake of conciseness - these are for the CIFAR-100 and CIFAR-10 datasets for models trained using each of the three primary losses. The plots demonstrate that TS-SB-ECE outperforms TS on the i.i.d. test set and under all levels of skew in the CIFAR-100-C and CIFAR-10-C datasets. Table \ref{tab:all-test-results} shows that TS-SB-ECE outperforms TS in 35 of the 42 settings that we experiment on. We conclude that Soft Calibration Objectives can be used to improve upon TS - the popular post-hoc calibration method.

A pertinent follow-up question for temperature scaling is whether we can take this approach of directly optimizing temperature for SB-ECE a step further and directly optimize temperature for (hard-binned) ECE instead. ECE is not trainable, but we might still be able to do a search for temperature. Indeed, multi-resolution search to optimize temperature for hard-binned ECE (hereby, TS-HB-ECE) is an alternative to gradient-based training of temperature for soft-binned ECE (i.e. TS-SB-ECE). Our results suggest that this might be promising. This approach is worth investigating further.

The two methods have potential advantages and disadvantages. TS-HB-ECE has higher variance than TS-SB-ECE for a given sample size, which risks a suboptimal solution even if TS-HB-ECE is the quantity we wish to optimize. TS-HB-ECE is also less computationally efficient than TS-SB-ECE. To formally compare complexities, assume that we want to compute top-label equal-width ECE and that we save logits in the last training epoch. Let $K$ be the number of classes, $N_c$ be size of the recalibration dataset and $V$ be the number of temperature values inspected for multi-resolution search. Assuming that the gradient-based method trains for a small constant number of epochs, the post-processing complexity for TS-SB-ECE is $O(N_cK)$ and for TS-HB-ECE is $O(VN_cK)$. The quantity $V$ may not be small if we want to get close to the optimal temperature. Consequently, this cost difference can be high for language models with large vocabularies and a lot of training data.

\begin{table}
\caption{Overcofident training set memorization happens in 2 of our 7 dataset + primary loss settings. This is characterized by a very low train ECE (in bold) and a high ratio of train ECE to test ECE (in bold). ECE is computed with the $\ell_1$ norm and equal-width binning. We use interleaved training with Soft Calibration Objectives as secondary losses in these 2 cases to reduce train ECE.}
\label{interleaving-table}
\vspace{0.5em}
  \centering
  \resizebox{0.8\textwidth}{!}{
  \begin{tabular}{||c|c||c||c|c||}
    \hline
    Dataset & Primary Loss & Test ECE & Train ECE & Test ECE / Train ECE \\
    \hline
    CIFAR-100 & NLL & 6.88\% & \textbf{0.45\%} & \textbf{15.14} \\
     & Focal & 4.21\% & 9.18\% & 0.46 \\
     & MSE & 3.51\% & 2.49\% & 1.41 \\
    \hline
    CIFAR-10 & NLL & 2.66\% & \textbf{0.05\%} & \textbf{49.82} \\
     & Focal & 5.28\% & 7.13\% & 0.74 \\
     & MSE & 6.53\% & 9.82\% & 0.66 \\
    \hline
    Imagenet & NLL & 3.35\% & 4.34\% & 0.77 \\
    \hline
  \end{tabular}
  }
\end{table} \hfill

\section{Reliability Plots}
\label{sec:reliability-plots}

\begin{figure}[bt!]
\centering
    \begin{subfigure}{\textwidth}
        \centering
        \includegraphics[width=\textwidth]{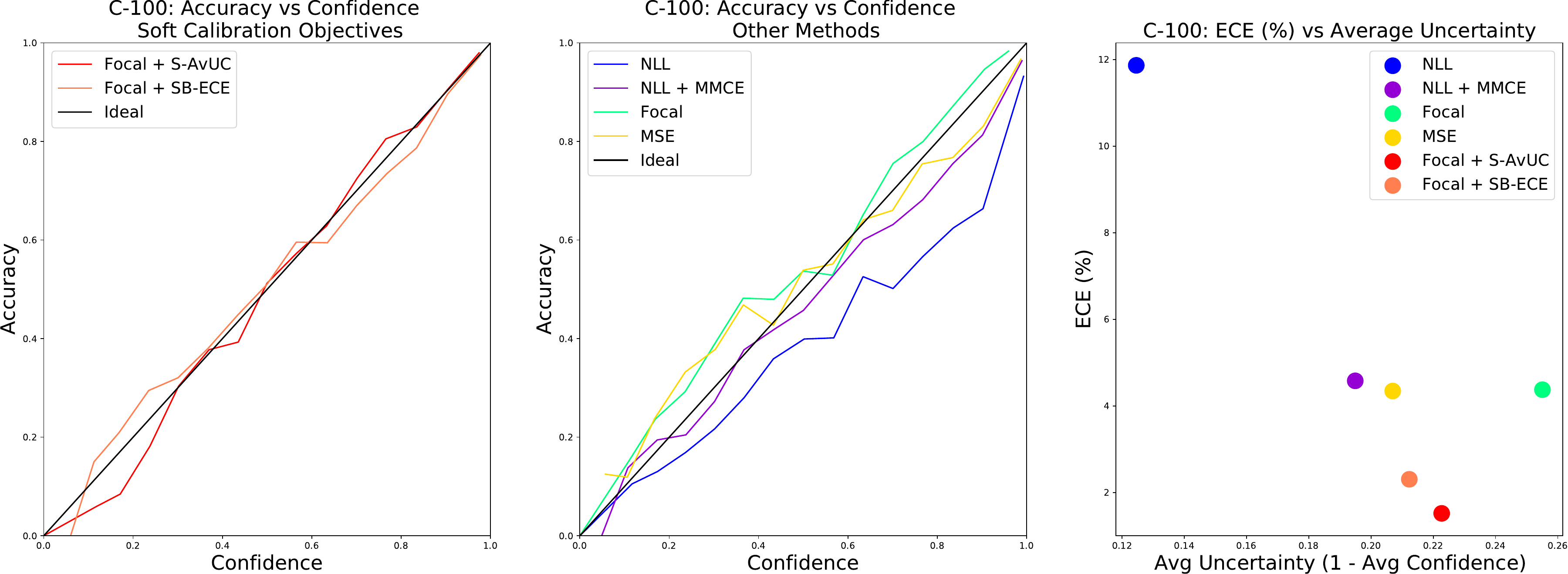}
        \subcaption{Reliability plots for C-100}
    \end{subfigure} \\
    \medskip
    \begin{subfigure}{\textwidth}
        \centering
        \includegraphics[width=\textwidth]{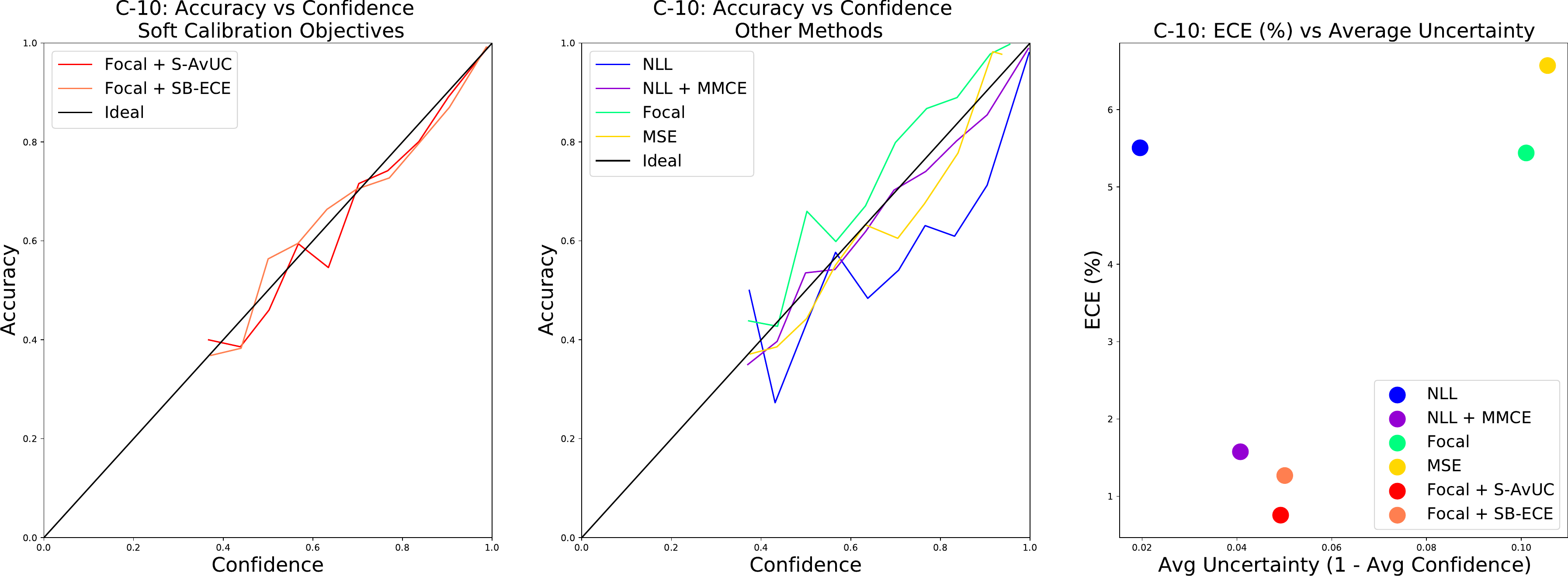}
        \subcaption{Reliability plots for C-10}
    \end{subfigure} \\
    \medskip
    \begin{subfigure}{\textwidth}
        \centering
        \includegraphics[width=\textwidth]{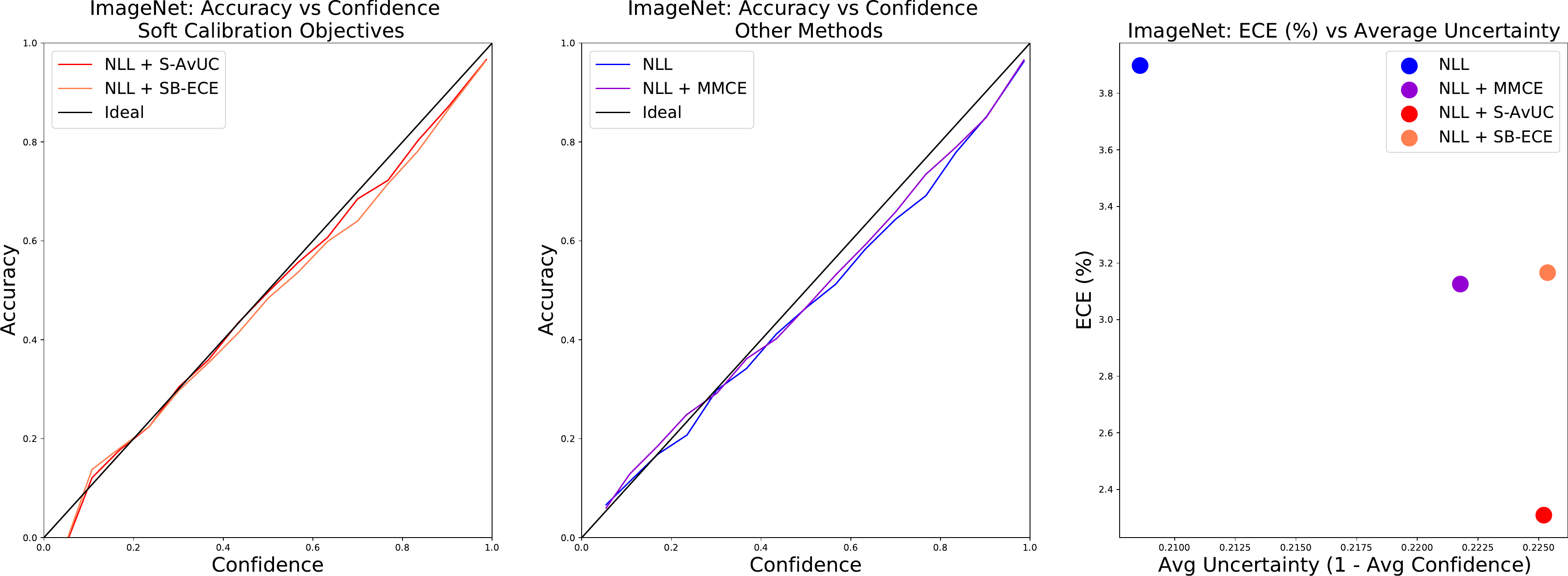}
        \subcaption{Reliability plots for ImageNet}
    \end{subfigure}
    \caption{Models trained with the \blue{NLL} loss are overconfident across all three datasets. Adding \purple{MMCE} as a secondary loss reduces the overconfidence. The \gold{MSE} loss results in underconfidence and overconfidence for different bins. Models trained with \green{Focal} loss are underconfident. Soft Calibration Objectives (\red{S-AvUC}, \orange{SB-ECE}) result in the most visually calibrated reliability plots across all datasets. Note that the high confidence regions have much higher density than the low confidence regions and are thus more critical to the ECE value. \\
    In the ECE vs average uncertainty plots, we see that \green{Focal} and \blue{NLL} losses result in the highest and lowest average uncertainties respectively. \red{S-AvUC} results in the lowest ECE in all three datasets and \orange{SB-ECE} is the next best for CIFAR-10 and CIFAR-100. Obtaining lower ECE as well as lower average uncertainty as compared to the \blue{NLL} loss remains an open challenge.}
    \label{fig:reliability}
\end{figure}

In this section, we use reliability plots (figure \ref{fig:reliability}) to compare the calibration of models trained using different training objectives for CIFAR-100, CIFAR-10 and ImageNet. We compare the cross-entropy baseline with each of the proposed methods to train for calibration: Focal loss \citep{lin2018focal}, MSE Loss \citep{hui2021evaluation}, MMCE loss \citep{Kumar2018TrainableCM}, S-AvUC and SB-ECE. We do not include the AvUC \citep{Krishnan2020ImprovingMC} loss since it does not help to improve calibration for non-Bayesian neural networks (section \ref{sec:avuc-gs}). We find that NLL results in overconfident models and that adding MMCE as a secondary loss to the NLL primary loss reduces the amount of overconfidence. The MSE primary loss results in models that are overconfident for some confidence bins and underconfident for others, which helps to explain why temperature scaling is not as effective for MSE as compared to Focal and NLL (see table \ref{tab:all-test-results}). Focal loss, on the other hand, results in underconfident models. This is also seen in the ECE vs average uncertainty plots (figure \ref{fig:reliability}), where we find that NLL results in least average uncertainty for all datasets whereas Focal loss is amongst the highest average uncertainties for both CIFAR-100 and CIFAR-10. Obtaining both lower ECE and lower average uncertainty simultaneously as compared to the standard cross-entropy loss remains an open challenge. Finally we note that models trained using Soft Calibration Objectives as secondary losses are the most visually calibrated for all three datasets, consistent with our findings in Section 5.



\section{Other Calibration Measures}
\label{sec:other-calibration-measures}

Even though ECE is the most popular metric for measuring calibration, it has issues related to consistency and bias \citep{nixon2019measuring,Kumar2019VerifiedUC,Roelofs2020MitigatingBI, Gupta2020CalibrationON}. Debiased CE \citep{Kumar2019VerifiedUC} and mean-sweep CE \citep{Roelofs2020MitigatingBI} have been shown to have lesser bias and more consistency across the number of bins parameter than ECE whereas KS-error \citep{Gupta2020CalibrationON} avoids binning altogether.

We have validated our findings on CIFAR-100 and CIFAR-10 using KS-error and mean-sweep CE. The findings based on these measures (table \ref{tab:other_calibration_measures_results}) are consistent with those based on ECE i.e. soft-calibration objectives outperform all other methods. We have reported results corresponding only to tables \ref{cifar100-table} and \ref{cifar10-table} for conciseness rather than those correponding to the full table \ref{tab:all-test-results}. As stated before, these account for the best-performing experiments on these datasets and in particular the best results for the cross-entropy baseline, Focal loss \citep{Mukhoti2020CalibratingDN} and MMCE \citep{Kumar2018TrainableCM}.

\begin{table}[h]
\caption{ We report accuracy, average ECE, KS-Error and mean-sweep CE for the CIFAR-10 and CIFAR-100 datasets corresponding to the entries in tables \ref{cifar100-table} and \ref{cifar10-table}. Mean-sweep CE is computed with the $\ell_2$ norm and equal-mass binning. KS-error is computed as originally defined with the $\ell_1$ norm. These additional measures further demonstrate that soft-calibration objectives outperform all other methods.}
\centering
\begin{subtable}[c]{0.8\textwidth}
  \subcaption{CIFAR-100}
  \centering
  \resizebox{\textwidth}{!}{
  \begin{tabular}{||c||c||c|c|c||}
    \hline
    Loss Fn. & Accuracy & ECE & KS-Error & mean-sweep CE \\
    \hline
    NLL & 78.7 & 9.10 & 6.71 & 9.02 \\
    \hline
    \makecell {NLL +  MMCE} & 77.2 & 4.77 & 3.27 & 4.77 \\
    \hline
    Focal & 78.3 & 4.66 & 4.20 & 4.57 \\
    \hline
    \makecell {Focal +  SB-ECE} & 78.6 & 2.30 & 0.71 & 2.21 \\
    \hline
    \makecell {Focal +  S-AvUC} & 78.1 & \textbf{1.57} & \textbf{0.51} & \textbf{1.24} \\
    \hline
  \end{tabular}
  }
\end{subtable} \hfill \\
\begin{subtable}[c]{0.8\textwidth}
\centering
  \subcaption{CIFAR-10}
  \resizebox{\textwidth}{!}{
  \begin{tabular}{||c||c||c|c|c||}
    \hline
    Loss Fn. & Accuracy & ECE & KS-Error & mean-sweep CE \\
    \hline
    NLL & 95.5 & 5.59 & 2.64 & 4.78 \\
    \hline
    \makecell {NLL +  MMCE} & 95.0 & 1.55 & 0.83 & 1.30 \\
    \hline
    Focal & 95.0 & 5.45 & 5.22 & 5.41 \\
    \hline
   \makecell {Focal +  SB-ECE} & 95.1 & \textbf{1.19} & \textbf{0.41} & \textbf{0.77} \\
    \hline
    \makecell {Focal + S-AvUC} & 94.4 & 1.58 & 0.66 & 1.43 \\
    \hline
  \end{tabular}
  }
\end{subtable}
\label{tab:other_calibration_measures_results}
\end{table}

\section{AvUC with Gradient Stopping}
\label{sec:avuc-gs}

The AvUC loss \citep{Krishnan2020ImprovingMC} was proposed to train for calibration in Stochastic Variational Inference (SVI) settings. It is based on the idea of giving an incentive to the model to be certain when accurate and uncertain when inaccurate via a secondary loss. The secondary loss term is described in equations \ref{eq:avuc} and \ref{eq:avuc-nxx}. These are restated here for readability.

\vspace{1ex}
\begin{align}\tag{13}
\textrm{AvUC}(\kappa, \hat D, \vtheta) = \log\Big(1 + \frac{n_\textrm{AU} + n_\textrm{IC}}{n_\textrm{AC} + n_\textrm{IU}}\Big),
\end{align}
\begin{equation}\tag{14}
  \begin{split}
    n_\textrm{AU} &= \Sigma_{i|(\vx_i, y_i) \in S_\textrm{AU}} (c_i \tanh h_i) \\
    n_\textrm{AC} &= \Sigma_{i|(\vx_i, y_i) \in S_\textrm{AC}} (c_i (1 - \tanh h_i))
  \end{split}
\quad\quad
  \begin{split}
    n_\textrm{IC} &= \Sigma_{i|(\vx_i, y_i) \in S_\textrm{IC}} ((1 - c_i) (1 - \tanh h_i)) \\
    n_\textrm{IU} &= \Sigma_{i|(\vx_i, y_i) \in S_\textrm{IU}} ((1 - c_i) \tanh h_i).
  \end{split}
\end{equation}

Note that $S_{AU}$, $S_{IC}$, $S_{AC}$ and $S_{IU}$ form a partition of datapoints from the training batch $\hat D$ which fall in each of the four categories resulting from two classifications: (1) accurate [A] vs. inaccurate [I] based on whether the model's prediction is correct and (2) certain [C] vs uncertain [U] whether the model's entropy is above or below a threshold $\kappa$. In our experiments we tune $\kappa$ rather than inferring it from the first few epochs as was suggested in \citep{Krishnan2020ImprovingMC}. Despite this additional degree of freedom, we consistently observe that the originally proposed AvUC loss does not help for calibration in non-Bayesian settings.

A closer look at equations \ref{eq:avuc} and \ref{eq:avuc-nxx} suggests that this might be because of some of the incentives provided by the secondary loss. As observed in section 4.2, minimizing the AvUC loss results in the model being incentivized to be even more confident in its inaccurate and certain predictions via minimizing $n_\textrm{IC}$, specifically the $(1 - c_i)$ multiplicand. Similarly, it also encourages the model to be even less confident in its accurate and uncertain predictions via minimizing $n_\textrm{AU}$, specifically the $c_i$ multiplicand.

To test whether these misincentives are really the cause of our observations, we conduct experiments with a variant of the loss where we stop gradients flowing through the $(1 - c_i)$ and $c_i$ multiplicands in each of the four expressions in equation \ref{eq:avuc-nxx}. We denote this modified version of the AvUC loss as the AvUC-GS loss, where "GS" denotes gradient stopping. The experiments confirm our hypothesis - we observe that the AvUC-GS secondary loss helps in improving calibration in situations where the original AvUC secondary loss did not. This observation holds across datasets and primary losses. This can be seen in table \ref{tab:all-test-results}, where the rows corresponding to AvUC-GS in the secondary loss column have lower ECE on an average than the respective rows corresponding to the AvUC secondary loss. We conclude that AvUC-GS is also an effective secondary loss. However, the soft calibration objective S-AvUC which is inspired from AvUC-GS generalizes, outperforms and supersedes it.

\section{Hyperparameter Tuning and the Accuracy-Calibration Tradeoff}
\label{sec:hyperparam-tuning}

We look at both the accuracy and the ECE of competing hyperparameter configurations. Some comparisons yield a clear winner but often there is a tradeoff between these. In such cases, we choose the lowest ECE whilst giving up less than 1\% accuracy relative to the hyperparameter configuration with the highest accuracy.

As stated in section 5.1, we use the Wide-Resnet-28-10 architecture \citep{zagoruyko2017wide} trained for 200 epochs on CIFAR-100 and CIFAR-10. For Imagenet, we use the Resnet-50 \citep{he2015deep} architecture trained for 90 epochs. The loss function we use in our experiments is $\textrm{PL} + \beta \cdot \textrm{SL} + \lambda \cdot \textrm{L2}$ where PL and SL denote the primary and secondary losses respectively and L2 denotes the weight normalization term with $\ell_2$ norm.

All our experiments use the SGD with momentum optimizer with momentum fixed to 0.9 and base learning rate fixed to 0.1. We follow a learning rate schedule which is fixed for each dataset across training losses. The number of bins ($M$, if applicable) for the soft binning secondary loss is fixed to 15. We used a per-core-batch-size of 64 on a 2x2 TPU topology with 8 cores for an effective batch size of 512. Both the baseline runs and experimental runs (all runs from Table \ref{tab:all-test-results}) used this batch size. In general, larger batch size is better for the computation of soft calibration losses, but we did not see significant ECE gains with a further increase in batch size. These form the set of hyperparameters we fixed rather than tuned. Fixing some hyperparamers allows us to keep the search space manageable. The values we use for these are those which work well for the cross-entropy baseline.

The $\beta$ (if applicable) and $\lambda$ parameters along with parameters relevant to the secondary loss are the set of hyperparameters that we tune. In our experiments with the SB-ECE secondary loss, the secondary loss function we use is $\textrm{SB-ECE}_{lb,p}(M, T, \hat D,\vtheta)$ from equation 12, for which we tune the $T$ parameter. In our experiments with the Soft-AvUC loss the secondary loss function we use is $\textrm{S-AvUC}(\kappa, T, \hat D, \vtheta)$ from equation 15, for which we tune the $\kappa$ and $T$ parameters. As stated above, these combined with $\beta$ and $\lambda$ form our set of tuned parameters. These hyperparameters are the most critical ones for demonstrating the effectiveness of our techniques.

We tune these parameters one-at-a-time starting with the threshold parameter for the secondary loss ($\kappa$, if applicable), followed by temperature parameter for the secondary loss ($T$, if applicable), the $\beta$ parameter (if applicable) and the L2-normalization coefficient $\lambda$. We retain the best value from the tuning experiments for one parameter while tuning a subsequent parameter.

We show in table \ref{tab:all-test-results} that Soft Calibration Objectives result in lower ECEs than previous methods in exchange for a small reduction in accuracy relative to the cross-entropy baseline. In our experiments, we found that other methods to train for calibration (MMCE, AvUC, Focal loss, MSE loss) also have to sacrifice a small amount of accuracy relative to the cross-entropy baseline in order to attain better calibrated models. This fundamental tradeoff can be summarized by the pareto-optimal curve between accuracy and calibration. Our methods result in points on this curve which are better calibrated than previously proposed methods, whilst trading off less than 1\% accuracy.

\section{Is SB-ECE as secondary loss a proper scoring rule?}
\label{sec:sb-ece-proper-loss}

We start this discussion by asking the following question: what happens to SB-ECE for perfectly calibrated models as the dataset size goes to infinity? We know that ECE tends to zero in this case. Proposition \ref{sb-ece-proposition} shows that the same holds for SB-ECE. We will use terminology introduced in sections \ref{sec:background} and \ref{sec:soft-calibration-objectives} in this section.

\begin{proposition}
\label{sb-ece-proposition}
Consider a dataset $D= \langle (\vx_i,y_i) \rangle _{i=1}^N$ drawn from the joint probability distribution $\mathcal D (\mathcal X, \mathcal Y)$. Say we have a perfectly calibrated model for it such that $E[a|c]=c$, where $a$ and $c$ denote accuracy and confidence respectively. If we consider $\textrm{SB-ECE}_{\textrm{bin},p}$ with $M$ bins, temperature $T$ and norm $p$ as defined in equation \ref{eq:ece-bin-soft}, we have
$$\textrm{SB-ECE}_{\textrm{bin},p}(M, T, \hat D,\vtheta) \rightarrow 0 \textrm{ as } N \rightarrow \infty$$
\end{proposition}

\begin{proof}
Let $p_c$ denote the p.d.f. of the confidence $c$ viewed as a random variable. Let us denote the membership function for bin $j$ by ${u_j}(c)$, a shorthand for our earlier notation ${u_{M,T,j}}(c)$ from section \ref{sec:sb-ece}. The size, confidence and accuracy of bin $j$, as defined in equations \ref{eq:bin-conf-soft} and \ref{eq:bin-acc-soft} are denoted by $C_j$ and $A_j$ respectively. We observe that as $N \rightarrow \infty$, these quantities satisfy the following:
\begin{align*}
C_j = \frac{\Sigma_{i=1}^{N} u_j(c_i) \cdot c_i}{\Sigma_{i=1}^{N} u_j(c_i)} &\rightarrow \frac{\int_0^1 xu_j(x)\,p_c(x)dx}{\int_0^1 u_j(x)\,p_c(x)dx} \\
A_j = \frac{\Sigma_{i=1}^{N} u_j(c_i) \cdot a_i}{\Sigma_{i=1}^{N} u_j(c_i)} &\rightarrow \frac{\int_0^1 E[a|c=x]u_j(x)\,p_c(x)dx}{\int_0^1 u_j(x)\,p_c(x)dx} \\
\intertext{Since $E[a|c=x]=x$ for perfectly calibrated models, we infer that as $N \rightarrow \infty$:}
C_j-A_j &\rightarrow 0 \\
\textrm{SB-ECE}_{\textrm{bin},p}(M, T, \hat D,\vtheta) &\rightarrow 0
\end{align*}
\end{proof}

Note that this does not necessarily hold for datasets of a given finite size. If we measure SB-ECE with $N=2$ datapoints for a perfectly-calibrated model which always has a confidence of $30\%$ and is correct $30\%$ of the time, we will always end up with ECE and SB-ECE both greater than zero.

If we consider an optimal classifier which always outputs the true probabilities, then it minimizes NLL since NLL is a proper scoring rule and it minimizes SB-ECE (note that SB-ECE $\geq 0$ by definition) as dataset size goes to infinity as per proposition \ref{sb-ece-proposition} since it is perfectly calibrated. Hence, it minimizes any positive linear combination of NLL and SB-ECE which implies that these linear combinations are proper scoring rules in the limit of infinite data.

\section{Post-hoc Dirichlet Calibration}
\label{sec:dirichlet-calibration}

In Tables \ref{tab:summary_results} and \ref{tab:all-test-results}, we have compared our methods to existing calibration-incentivizing losses that operate during training, with and without post-hoc temperature scaling. There are a several post-hoc recalibration techniques which can be used complementary to our methods. For this reason, comparing to these has not been the focus of our work, similar to \citep{Mukhoti2020CalibratingDN} and \citep{Kumar2018TrainableCM}. Nevertheless, in this section we show that our methods do significantly better than training with the cross-entropy, focal or MSE primary losses and using such post-hoc recalibration methods. In particular, we compare against Dirichlet calibration \citep{Kull2019BeyondTS}. Table \ref{tab:dirichlet-table} records ECE numbers for the best-performing training objectives for CIFAR-10 corresponding to table \ref{cifar10-table}, both with and without post-hoc Dirichlet calibration.

\begin{table}[h]
\caption{We report accuracy and average ECE across 10 runs for the best performing training objectives for the CIFAR-10 dataset corresponding to the entries in table \ref{cifar10-table}, both with and without post-hoc dirichlet calibration. ECE is computed with the $\ell_2$ norm and equal-mass binning. Our methods outperform post-hoc dirichlet calibration applied to the NLL, Focal and MSE primary losses.}
\centering
\begin{subtable}[c]{0.65\textwidth}
\centering
  \subcaption{CIFAR-10}
  \resizebox{\textwidth}{!}{
  \begin{tabular}{||c||c||c|c||}
    \hline
    Loss Fn. & Accuracy & ECE & ECE w/ Dirichlet \\
    \hline
    NLL & 95.5 & 5.59 & 2.45 \\
    \hline
    \makecell {NLL +  MMCE} & 95.0 & 1.55 & \textbf{1.28} \\
    \hline
    Focal & 95.0 & 5.45 & 2.48 \\
    \hline
   \makecell {Focal +  SB-ECE} & 95.1 & \textbf{1.19} & 2.24 \\
    \hline
    \makecell {Focal + S-AvUC} & 94.4 & 1.58 & 1.70 \\
    \hline
    MSE & 95.0 & 6.58 & 4.69 \\
    \hline
  \end{tabular}
  }
\end{subtable}
\label{tab:dirichlet-table}
\end{table}

\pagebreak

\section{Multiclass Calibration Error}
\label{sec:multiclass-calibration}

Reducing top-label calibration error using calibration-incentivizing training loss functions is the focus of our work, similar to \citep{Kumar2018TrainableCM}, \citep{Krishnan2020ImprovingMC} and \citep{Mukhoti2020CalibratingDN}. In this section, we additionally evaluate marginal ECE \citep{Kumar2019VerifiedUC} which measures the calibration of the predicted probability distribution over all classes rather than just the top class. Table \ref{tab:multiclass-ce} shows that our methods which are trained to minimize top-label ECE yield the best marginal ECE numbers as well.

The terms in the S-AvUC loss are tied to the top-label and this does readily lend itself to a multiclass extension. The soft binning approach however immediately yields a trainable soft version of marginal ECE analogous to the top-label case. Soft-binned marginal ECE is beyond the scope of this paper but can be investigated further.

\begin{table}[h]
\caption{We report accuracy, average top-label ECE and average marginal ECE for the CIFAR-10 and CIFAR-100 datasets corresponding to the entries in tables \ref{cifar100-table} and \ref{cifar10-table}. Marginal ECE and top-label ECE are computed with the $\ell_2$ norm and equal-mass binning. Soft-calibration objectives outperform other methods on multiclass CE in addition to the top-label CE they are designed to optimize.}
\centering
\begin{subtable}[c]{0.65\textwidth}
  \subcaption{CIFAR-100}
  \centering
  \resizebox{\textwidth}{!}{
  \begin{tabular}{||c||c||c|c||}
    \hline
    Loss Fn. & Accuracy & ECE & Marginal ECE \\
    \hline
    NLL & 78.7 & 9.10 & 3.78 \\
    \hline
    \makecell {NLL +  MMCE} & 77.2 & 4.77 & 3.05 \\
    \hline
    Focal & 78.3 & 4.66 & 4.35 \\
    \hline
    \makecell {Focal +  SB-ECE} & 78.6 & 2.30 & 3.05 \\
    \hline
    \makecell {Focal +  S-AvUC} & 78.1 & \textbf{1.57} & \textbf{2.72} \\
    \hline
  \end{tabular}
  }
\end{subtable} \hfill \\
\begin{subtable}[c]{0.65\textwidth}
\centering
  \subcaption{CIFAR-10}
  \resizebox{\textwidth}{!}{
  \begin{tabular}{||c||c||c|c||}
    \hline
    Loss Fn. & Accuracy & ECE & Marginal ECE \\
    \hline
    NLL & 95.5 & 5.59 & 2.31 \\
    \hline
    \makecell {NLL +  MMCE} & 95.0 & 1.55 & 2.06 \\
    \hline
    Focal & 95.0 & 5.45 & 5.64 \\
    \hline
   \makecell {Focal +  SB-ECE} & 95.1 & \textbf{1.19} & 2.24 \\
    \hline
    \makecell {Focal + S-AvUC} & 94.4 & 1.58 & \textbf{1.65} \\
    \hline
  \end{tabular}
  }
\end{subtable}
\label{tab:multiclass-ce}
\end{table}

\end{document}

%% file: math_commands.tex
\usepackage{amsmath,amsfonts,bm}

\def\eqref#1{equation~\ref{#1}}

\def\1{\bm{1}}

\def\vtheta{{\bm{\theta}}}

\def\vf{{\bm{f}}}
\def\vg{{\bm{g}}}

\def\vu{{\bm{u}}}
\def\vv{{\bm{v}}}

\def\vx{{\bm{x}}}

\DeclareMathAlphabet{\mathsfit}{\encodingdefault}{\sfdefault}{m}{sl}
\SetMathAlphabet{\mathsfit}{bold}{\encodingdefault}{\sfdefault}{bx}{n}

\newcommand{\softmax}{\mathrm{softmax}}

